\documentclass{article}
\usepackage{microtype}
\usepackage{graphicx}
\usepackage{booktabs}
\usepackage{hyperref}
\usepackage{xcolor}

\usepackage{amsmath}
\usepackage{amsthm}
\usepackage{amsfonts}
\usepackage{amssymb}
\usepackage{bbm}
\usepackage{multirow}
\usepackage[utf8]{inputenc}
\usepackage{tikz}
\usepackage{verbatim}
\usepackage{listings}
\usepackage{subfig}
\usepackage{enumitem}
\usepackage[accepted]{icml2021}

\icmltitlerunning{Benchmarks, Algorithms, and Metrics for Hierarchical Disentanglement}

\begin{document}

\twocolumn[
\icmltitle{Benchmarks, Algorithms, and Metrics for Hierarchical Disentanglement}
\begin{icmlauthorlist}
\icmlauthor{Andrew Slavin Ross}{ha}
\icmlauthor{Finale Doshi-Velez}{ha}
\end{icmlauthorlist}
\icmlaffiliation{ha}{Harvard University, Cambridge, MA, USA}
\icmlcorrespondingauthor{Andrew Slavin Ross}{andrewslavinross@gmail.com}
\icmlkeywords{Machine Learning, ICML}
\vskip 0.3in
]

\printAffiliationsAndNotice{}

\begin{abstract}
In representation learning, there has been recent interest in developing algorithms to disentangle the ground-truth generative factors behind a dataset, and metrics to quantify how fully this occurs.  However, these algorithms and metrics often assume that both representations and ground-truth factors are flat, continuous, and factorized, whereas many real-world generative processes involve rich hierarchical structure, mixtures of discrete and continuous variables with dependence between them, and even varying intrinsic dimensionality.  In this work, we develop benchmarks, algorithms, and metrics for learning such hierarchical representations.
\end{abstract}

\section{Introduction}

% Motivation: we want to learn representations that model real life

Autoencoders aim to learn structure in data by compressing it to a lower-dimensional representation with minimal loss of information. Although they have proven useful in many applications~\citep{lecun2015deep}, the individual dimensions of their representations are often inscrutable, even when the underlying data is generated by simple processes. Motivated by needs for interpretability~\citep{alvarez2018towards,marx2019disentangling}, fairness~\citep{creager2019flexibly}, and generalizability~\citep{bengio2013representation}, as well as a basic intuition that representations should model the data correctly, a subfield has emerged which applies representation learning algorithms to synthetic datasets and checks how well representation dimensions ``disentangle'' the known ground-truth factors behind the dataset.

% Existing approach: disentangled representations. But these have problems, and attempts to solve them have been piecemeal.

Perhaps the most common disentanglement approach has been to learn continuous vector representations whose dimensions are statistically independent (and evaluate them using metrics that assume ground-truth factors are also independent), reasoning that factorization is a useful proxy \citep{ridgeway2016survey,higgins2017beta,chen2018isolating,kim2018disentangling}. However, this problem is not identifiable~\citep{locatello2018challenging}, and it seems unlikely that continuous, factorized, fixed-dimensional representations are the optimal choice for modeling many real-world generative processes, which are often highly structured, with nested parameters that only become active in particular cases.

As a concrete example, consider the problem of learning representations of medical phenotypes of patients with and without diabetes mellitus, a complex disease with multiple types and subtypes~\citep{american2005diagnosis}. Some underlying factors of phenotype variation---as well as the intrinsic complexity of these variations---are likely specific to the disease, its types, or its subtypes~\citep{ahlqvist2018novel}.
A representation that faithfully modeled the true factors of variation would need to be deeply hierarchical, with some dimensions only active for certain subtypes.
%But existing disentanglement methods
Ideally, it also should also be possible to learn such representations even if these subtypes (and the number of dimensions needed to parameterize them) are unknown. 

%However, this notion of an ``activity'' hierarchy differs from that most commonly considered in prior works on hierarchical representations, which focuses on conditional dependence~\cite{esmaeili2018structured,sonderby_ladder_2016}, and even those that do focus on shallow hierarchies that must be prespecified in advance.

%However, when ``hierarchies'' are considered in current disentanglement research, they generally do not involve activating or deactivating dimensions, and they also tend to be shallow and fully specified in advance.

Our approach in this paper is ambitious: we introduce (1) a flexible framework for modeling deep hierarchical structure in datasets, (2) novel algorithms for learning both structure and structured autoencoders entirely from data, which we apply to (3) novel benchmark datasets, and evaluate with (4) novel hierarchical disentanglement metrics.
Our framework is based on the idea that data may lie on multiple manifolds with different intrinsic dimensionalities, and that certain (nested groups of) dimensions may only be active for a subset of the data.
Though at first glance this approach seems it should worsen, not improve, identifiability, our structure assumptions also serve as an inductive bias that empirically help us learn representations that more faithfully (and explicitly) model ground-truth generative processes.

\section{Related Work}

In this section, we review work related to our notion of ``hierarchical disentangled representations.'' However, there are many notions of hierarchy that can be introduced into representations (or into definitions of disentanglement), some of which have little in common except a shift in focus away from flatness or factorization.

Still, the problem of learning a flat, factorized representation has received significant attention over the years. Much of the initial work, e.g. from \citet{schmidhuber1992learning}, \citet{zemel1994minimum}, and \citet{comon1994independent}, was motivated by classic problems like source separation or biological and information-theoretic arguments about minimum description length~\citep{barlow1961possible}. More recently, \citet{ridgeway2016survey} argued that factorization was often a useful real-world proxy for disentanglement in the seminal sense of \citet{bengio2013deep}, which motivated the development of a number of popular methods for training variational autoencoders (VAEs, \citet{kingma2013auto}) to reconstruct data from compressed flat vectors, but with minimal total correlation (TC) between their components~\citep{higgins2017beta,chen2018isolating,kim2018disentangling,dupont2018learning,kim2019relevance,jeong2019learning}.
We build on these approaches in our work, which we also tie back to some of their original motivating problems like minimum description length (see \S\ref{sec:mdl}).

There are, however, a number of limitations to learning factorized representations.
To begin with, the problem was actually shown by \citet{locatello2018challenging} to be non-identifiable, at least without weak supervision~\citep{locatello2020weakly,klindt2021towards}. More pressingly, though, factorization sometimes \emph{prevents} us from learning representations that disentangle independent causal mechanisms with nontrivial structure~\citep{parascandolo2018learning,trauble2020independence}, which is actually how \citet{bengio2013deep} defined the challenge of disentanglement. Our goal in this work is to learn representations that can identify and explicitly model this kind of structure when it exists.

Still, there are a wide variety of ways to incorporate structure into representations or disentanglement. One is simply to change the disentanglement objective, e.g. to encourage different degrees of factorization within and across subgroups~\citep{esmaeili2018structured}. Another is to change the representation architecture such that ``low-level'' components are drawn conditionally on ``high-level'' components from some fixed hierarchy or graphical model~\citep{sonderby_ladder_2016,siddharth2017learning,singh2019finegan}. Others use mixed discrete-continuous representations where continuous representation components are either ``global'' (marginally independent) or ``local'' to a specific categorical value (conditionally independent, and sometimes inactive when the categorical takes other values)~\citep{sorrenson2020disentanglement,choi2020discond}.
Typically, though, these approaches only support shallow hierarchies that must be specified by the user in advance, or require instance-level supervision~\citep{yang2020causalvae}.
Our work is closest to the global-local approach of \citet{choi2020discond}, but we support arbitrarily deep hierarchies, and also learn them from data.

%In this work, we return to the problem of learning disentangled representations from data alone. As our inductive bias to reduce (though not eliminate) non-identifiability, we assume the data contains discrete hierarchical structure that can be inferred geometrically. 
%Though this introduces challenging new problems, it also creates opportunities to learn interpretable global summaries of the data.

Other related approaches include relational autoencoders~\citep{wang2014generalized}, which model structure between non-iid flat data, and graph neural networks~\citep{defferrard2016convolutional}, which learn flat representations of structured data. In contrast, we model structure \emph{within} flat inputs.
Also relevant are advances in object representations, such as slot attention~\citep{locatello2020object}. While this area has generally not focused on hierarchical nesting, it learns structure and seamlessly handles sets; we view our method as complementary. Finally, our hierarchy detection method is built on work in multiple- and robust manifold learning~\citep{mahapatra2017s,mahler2020contagion}. We contribute new robustness innovations and also introduce hierarchy (like \citet{tino2002hierarchical}, but without requiring fixed dimensionality or human feedback).

\begin{figure}
    \centering
    \includegraphics[width=0.9\linewidth]{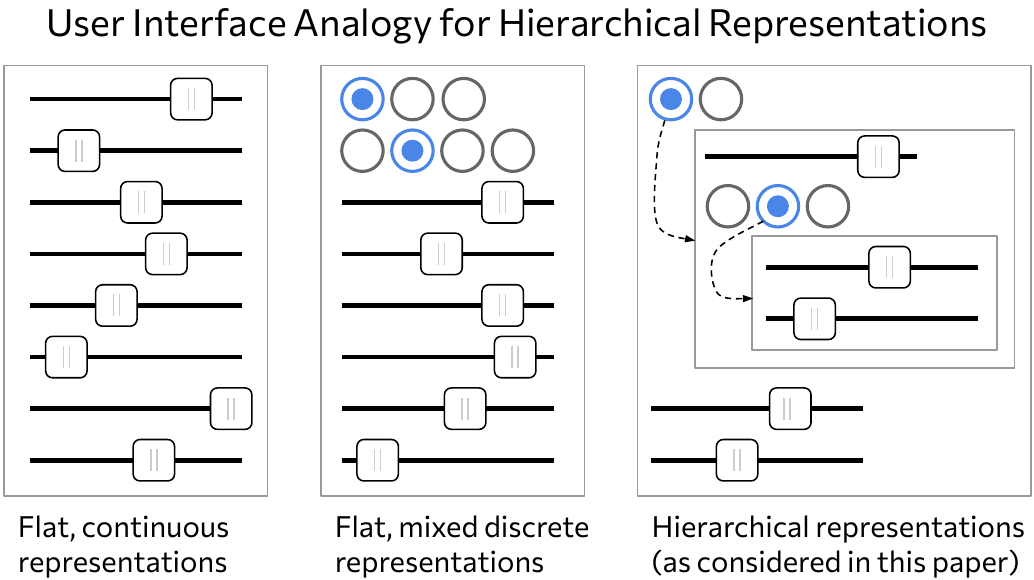}
    \caption{User interface analogy for our model of representation hierarchy. In traditional flat representations, all dimensions are active simultaneously, no matter whether they are continuous (shown as sliders) or discrete (shown as radio buttons). In our model, groups of dimensions may only be active for specific values of discrete dimensions, and groups can be nested.}
    \label{fig:hier-reps-html}
\end{figure}

\begin{figure*}[ht]
    \centering
    %\includegraphics[width=0.33\textwidth]{figures/spaceshapes-2x9.png}
    %\includegraphics[width=0.33\textwidth]{figures/chopsticks-222-2x9.png}
    %\includegraphics[width=0.33\textwidth]{figures/chopsticks-311-2x9.png}
    %%
     %\resizebox{\textwidth}{!}{
     
     %\noindent\begin{minipage}{\textwidth}
    
    \noindent\begin{minipage}[b]{.33\textwidth}\centering
     \includegraphics[width=\textwidth]{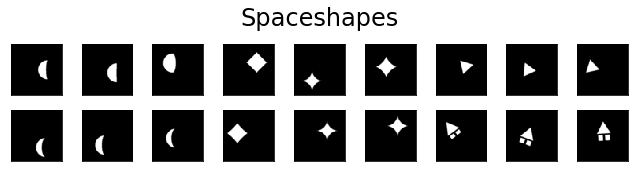}
    \includegraphics[height=1.1in,trim={1.5cm 1.33cm 1.5cm 1.33cm},clip]{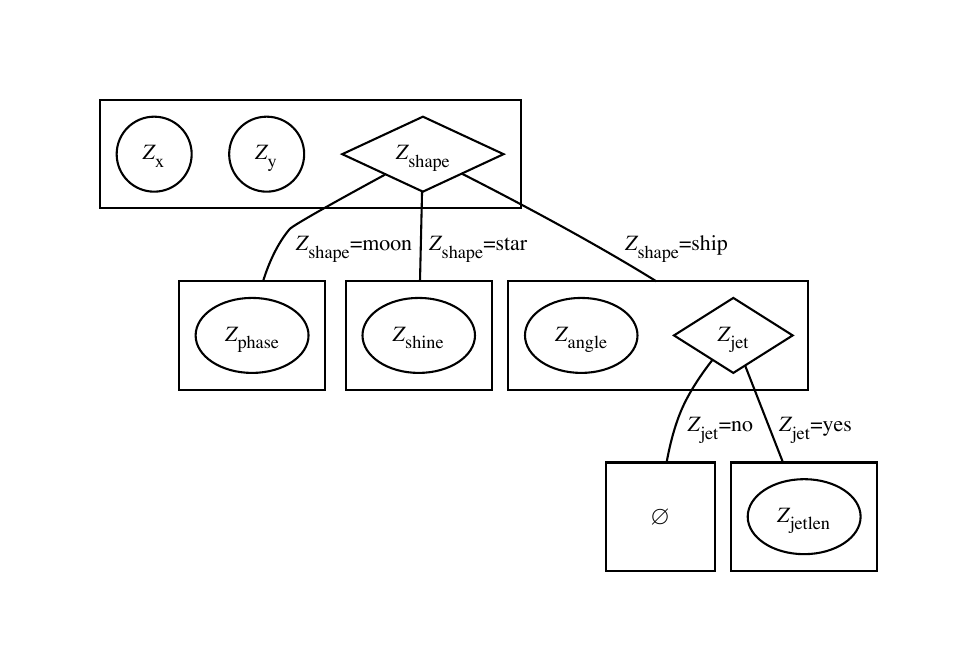}
     \end{minipage}
     \noindent\begin{minipage}[b]{.33\textwidth}\centering
     \includegraphics[width=\textwidth]{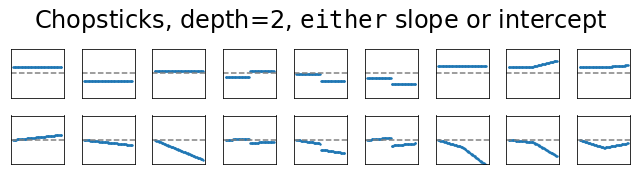}
     \includegraphics[height=1.1in,trim={1.5cm 1.33cm 1.5cm 1.33cm},clip]{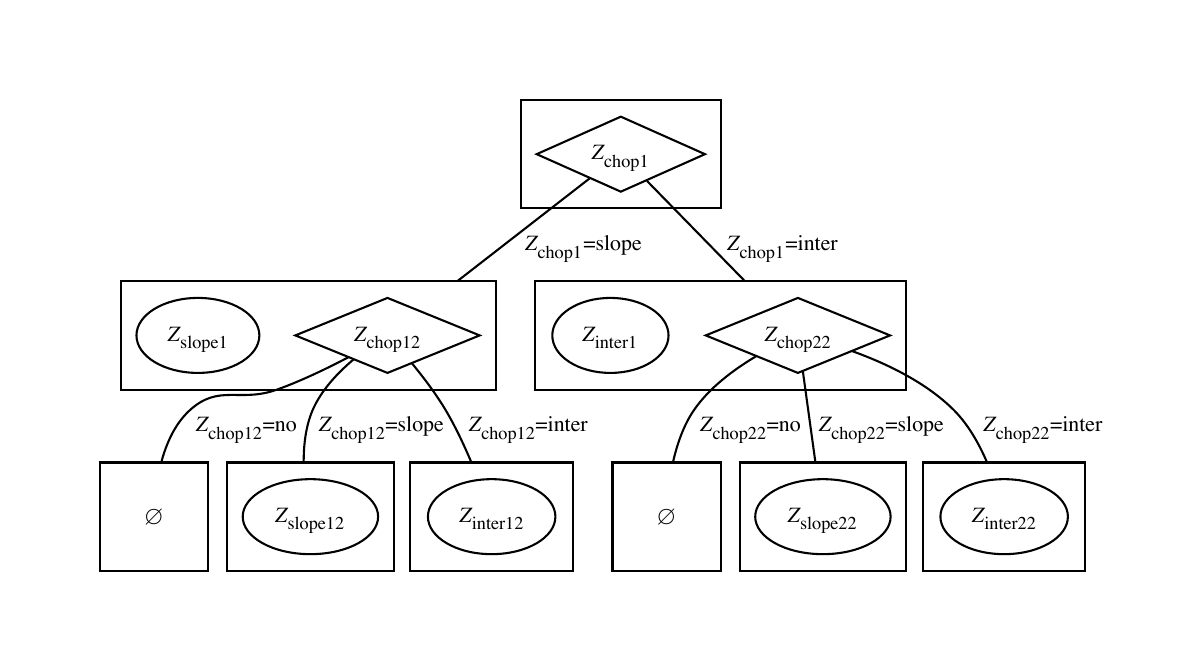}
     \end{minipage}
     \noindent\begin{minipage}[b]{.33\textwidth}\centering
    \includegraphics[width=\textwidth]{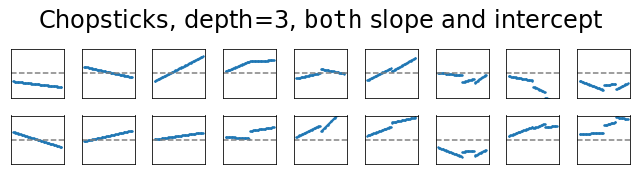}
    \includegraphics[height=1.1in,trim={1.5cm 1.33cm 1.5cm 1.33cm},clip]{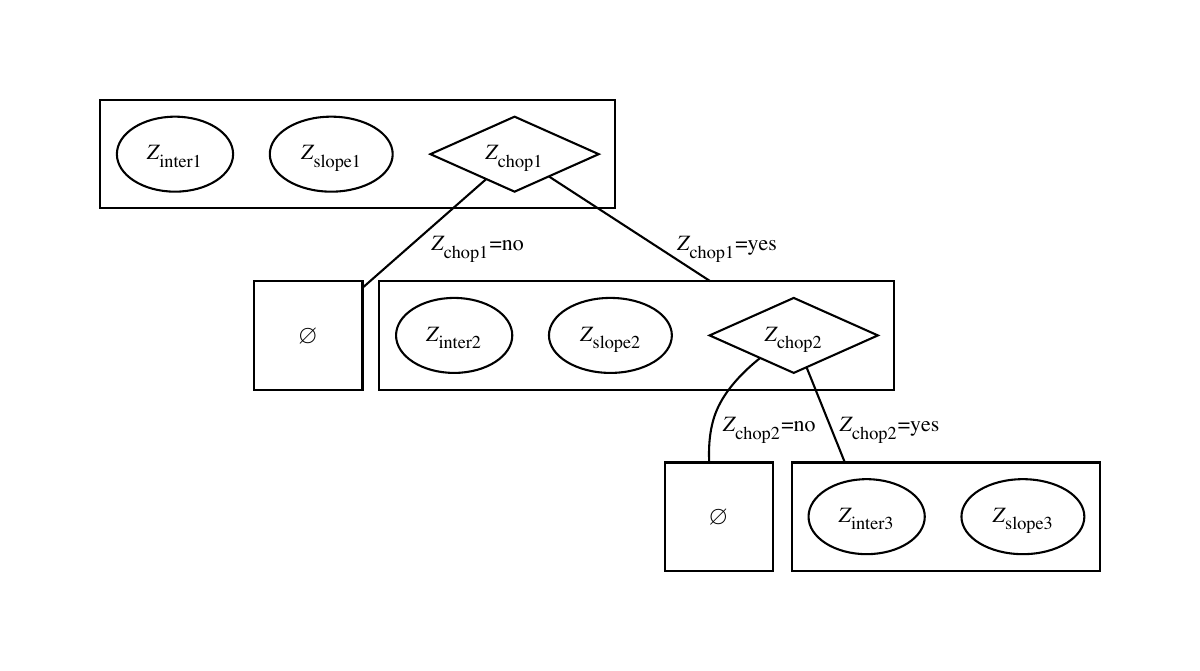}
    \end{minipage}
     
    % \end{minipage}
    
    %\hspace{3cm}

    \caption{Examples and ground-truth variable hierarchies for Spaceshapes and two different variants of Chopsticks. Continuous variables are shown as circles and discrete variables are shown as diamonds. Discrete variables have subhierarchies of additional variables that are only active for particular discrete values. See also Fig.~\ref{fig:hier-reps-html}-style interactive visualizations of hierarchical representations explicitly trained to match ground-truth for \href{https://hreps.s3.amazonaws.com/viz/index.html?dataset=spaceshapes&model=supervised}{each} \href{https://hreps.s3.amazonaws.com/viz/index.html?dataset=chopsticks_depth2_either&model=supervised}{dataset} \href{https://hreps.s3.amazonaws.com/viz/index.html?dataset=chopsticks_depth3_both&model=supervised}{respectively}.}
    \label{fig:dataset-examples}
\end{figure*}

\section{Hierarchical Disentanglement Framework}
In this section, we outline our framework for modeling hierarchical structure in representations.
In our framework, we associate individual data points with paths down a \emph{dimension hierarchy} (examples in Fig.~\ref{fig:dataset-examples}). Dimension hierarchies consist of dimension group nodes (shown as boxes), each of which can have any number of continuous dimensions (shown as ovals) and an optional categorical variable (diamonds) that leads to other groups based on its value. For any data point, we ``activate'' only the dimensions along its corresponding path. Notation-wise, $\mathtt{root}(Z)$ denotes the group at the root of a hierarchy, and $\mathtt{children}(Z_j)$ denotes the child groups of a categorical dimension $Z_j$. In the context of a dataset, for a dimension $Z_j$ or a dimension group $g$, $\mathtt{on}(Z_j)$ or $\mathtt{on}(g)$ denotes the subset of the dataset where that $Z_j$ or $g$ is active.

As a potentially more intuitive analogy (as well as a visualization method), we can also understand hierarchical representations in terms of user interfaces with nested groups of sliders and radio buttons (Fig.~\ref{fig:hier-reps-html}). While traditional representations might consist of a single group of constantly visible sliders (or a mixture of sliders and radio buttons), hierarchical representations contain subgroups that only appear when parent radio buttons take particular values. For such representations, only a subset of dimensions need to be visible to users at any given time, even if many are required to model the dataset---which could significantly improve interpretability \citep{ross2021evaluating}, if the hierarchy itself is comprehensible.

% Our motivation is interpretability. Preliminary studies suggest that, while disentanglement methods can significantly improve the interpretability of relatively low-dimensional representations, high-dimensional representations may be inherently hard to understand. However, if a representation could be structured such that only a subset of dimensions is relevant for most inputs

Although in this work we only consider tree-structured hierarchies, our framework could be extended to support multiple categoricals per node or even DAGs, such that instances can be associated with directed flows down multiple paths.

\section{Hierarchical Disentanglement Benchmarks}
\label{sec:benchmarks}

In this section, both to clarify our framework and enable testing of our algorithms, we introduce several synthetic benchmark datasets with ground-truth hierarchical structure (see Fig.~\ref{fig:dataset-examples} for instances and dimension hierarchies).
%For new frameworks, it is critical to have synthetic benchmarks for which the true structure is known and ground truth disentanglement scores can be computed. We present several in Fig.~\ref{fig:dataset-examples} and describe them below: 

\subsection{Spaceshapes}\label{sec:spaceshapes}
Our first benchmark dataset is Spaceshapes, a binary 64x64 image dataset meant to hierarchically extend dSprites, a shape dataset common in the disentanglement literature \citep{dsprites17}.  
Like dSprites, Spaceshapes images have location variables $\mathtt{x}$ and $\mathtt{y}$, as well as a categorical $\mathtt{shape}$ with three options (in our case, $\mathtt{moon}$, $\mathtt{star}$, and $\mathtt{ship}$). However, depending on shape, we add other continuous variables with qualitatively different effects: $\mathtt{moon}$s have a $\mathtt{phase}$; $\mathtt{star}$s have a sharpness to their $\mathtt{shine}$; and $\mathtt{ship}$s have an $\mathtt{angle}$. 
Finally, $\mathtt{ship}$s can optionally have a $\mathtt{jet}$, which has a length ($\mathtt{jetlen}$), but this is only defined at the deepest level of the hierarchy.  The presence of $\mathtt{jetlen}$ alters the intrinsic dimensionality of the representation; it can be either 3D or 4D depending on the path.
As in dSprites, variables are sampled from continuous or discrete uniforms. An interactive visualization of a representation trained to model this ground-truth hierarchy can be viewed \href{https://hreps.s3.amazonaws.com/viz/index.html?dataset=spaceshapes&model=supervised}{here}.

%The dimension hierarchy we present in Fig.~\ref{fig:dataset-examples} is arguably not the only reasonable choice; for example, $\mathtt{shape}$ sub-dimensions could all be modeled by a single variable at the top level, or $\mathtt{x}$ and $\mathtt{y}$ could instead be duplicated below each shape (see Fig.~\ref{fig:spaceshapes-alts} for alternate versions). However, each of those modifications preserves the overall structure of the hierarchy. We discuss how we handle this apparent ambiguity in Sec.~\ref{sec:metric-desiderata}

%Spaceshapes is also designed with a certain ambiguity in the dimension hierarchy; it is possible to faithfully model
%Note that there are arguably multiple dimension hierarchies that could reasonably characterize the Spaceshapes dataset, though all of them share a common structure.

\subsection{Chopsticks}

Our second benchmark, Chopsticks, is actually a family of arbitrarily deep timeseries datasets.
Chopsticks samples are 64D linear segments, each of which can have a uniform-sampled $\mathtt{slope}$ and/or $\mathtt{intercept}$; different dataset variants can have one, the other, $\mathtt{both}$, or $\mathtt{either}$ but not both.
For all variants, segments initially span the whole interval. 
However, we then flip a coin to determine whether to $\mathtt{chop}$ the segment, in which case we add a uniform offset to the slope and/or intercept of the right half.  We repeat this process recursively up to a configurable maximum $\mathtt{depth}$, biasing probabilities so that we have equal probability of stopping at each level; each chop requires increasing local dimensionality to track additional slopes and intercepts.
Although the underlying process is quite simple, the structure can be made arbitrarily deep, making it a useful benchmark for testing structure learning. 
We provide more details in \S\ref{sec:more-chopsticks}, and
interactive visualizations are also available for the \href{https://hreps.s3.amazonaws.com/viz/index.html?dataset=chopsticks_depth2_either&model=supervised}{depth-2 $\mathtt{either}$} and \href{https://hreps.s3.amazonaws.com/viz/index.html?dataset=chopsticks_depth3_both&model=supervised}{depth-3 $\mathtt{both}$} variants.
%Note that in the case of the depth-2 $\mathtt{either}$ variant of Chopsticks, the slope and intercept of the second half of a split segment are represented by different continuous variables based on whether the first half has variable slope or intercept.

Although these datasets are designed to have clear hierarchical structure, 
there are certain ambiguities in how to structure aspects of the dimension hierarchies, which we discuss in \S\ref{sec:metric-desiderata}.

%Although Note that in the case of the depth-2 $\mathtt{either}$ variant of Chopsticks, the slope and intercept of the second half of a split segment are represented by different continuous variables based on 

\section{Hierarchical Disentanglement Algorithms}

We next present a method for learning hierarchical disentangled representations from data alone. We split the problem into two brunch-themed algorithms, MIMOSA (which infers hierarchies) and COFHAE (which trains autoencoders).

\subsection{Learning Hierarchies with MIMOSA}

\begin{algorithm*}
    \caption{MIMOSA($X$)}
    \label{algo:mimosa}
\begin{algorithmic}[1]
    \STATE Encode the data $X$ using a smooth autoencoder to reduce the initial dimensionality. Store as $Z$.
    \STATE Construct a neighborhood graph on $Z$ using a Ball Tree \citep{omohundro1989five}.
    \STATE Run LocalSVD (Algorithm~\ref{algo:local-svd}) on each point in $Z$ and its neighbors to identify local manifold directions.
    \STATE Run BuildComponent (Algorithm~\ref{algo:build-component}) to successively merge neighboring points with similar local manifold directions.
    \STATE Run MergeComponents (Algorithm~\ref{algo:merge-components}) to combine similar components over longer distances and discard outliers.
    \STATE Run ConstructHierarchy (Algorithm~\ref{algo:construct-hier}) to create a dimension hierarchy based on which components enclose others.
    \STATE \textbf{return} the hierarchy and component assignments.
\end{algorithmic}
\end{algorithm*}

\begin{figure*}[h!]
    %\centering
    %\includegraphics[width=0.275\textwidth]{figures/chopsticks222-init-rep.png}
    %\includegraphics[width=0.385\textwidth]{figures/chopsticks222-learned-comps.png}
    %\includegraphics[width=0.3\textwidth]{figures/chopsticks_depth2_slope2_inter2_hier.png}
        \resizebox{\textwidth}{!}{
    \fbox{\includegraphics[height=3in]{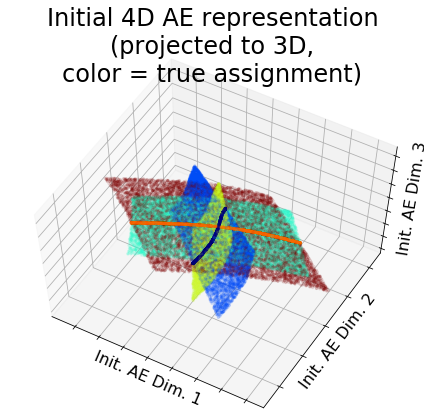}}
    \fbox{\includegraphics[height=3in]{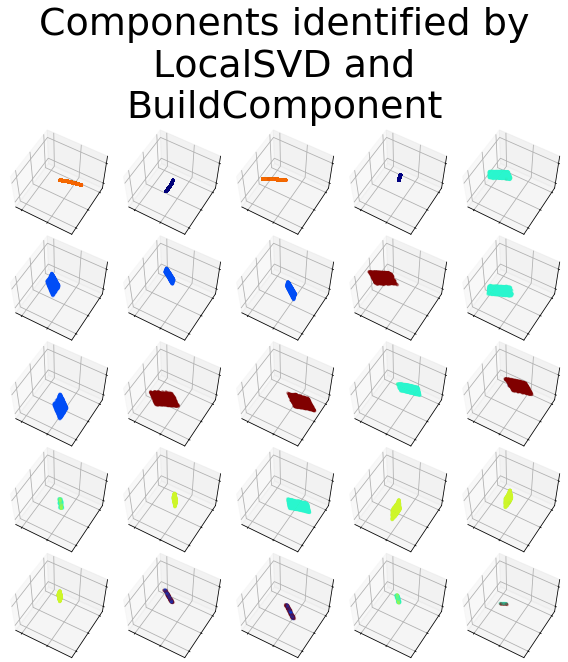}}
    \fbox{\includegraphics[height=3in]{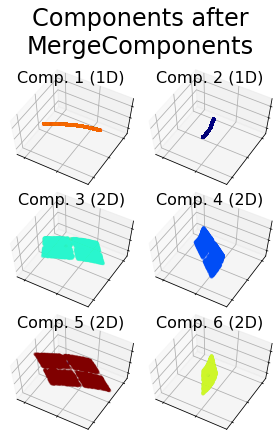}}
    \fbox{
    \begin{tikzpicture}
    %\node[inner sep=0] (image) at (0,0) {\includegraphics[height=3in,trim={1.5cm 1.5cm 1.5cm 1.5cm},clip]{figures/chopsticks222_pdf.pdf}};
    %\node[above=2cm of image] {Hello world};
    \tikzset{every node}=[font=\Large\sffamily]
      \draw (0, 0) node[inner sep=0] {\includegraphics[height=3in,trim={1.5cm 1.33cm 1.5cm 0.25cm},clip]{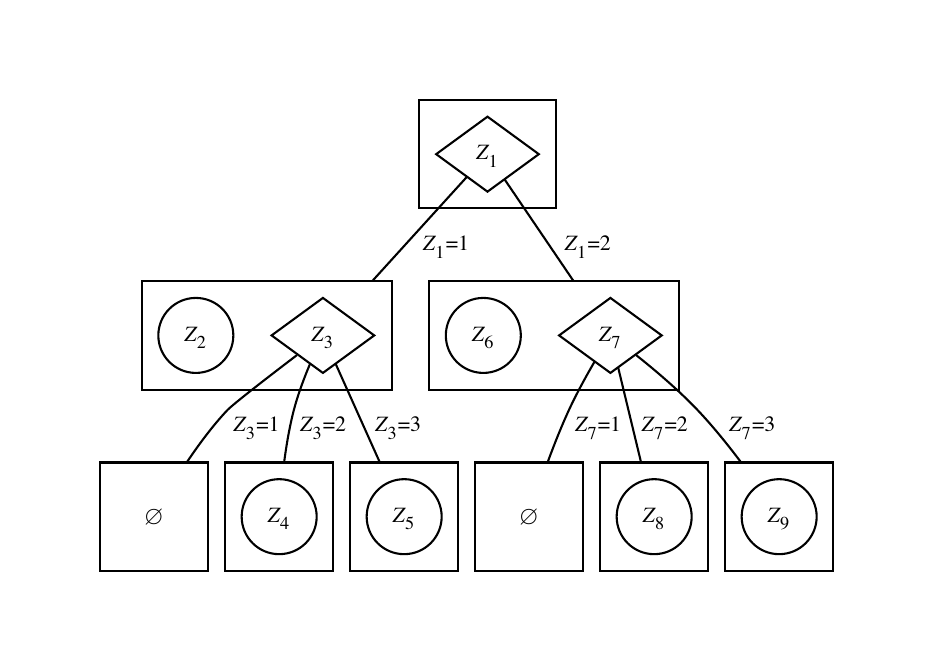}};
    \draw (0, 3.3) node {Result from ConstructHierarchy};
    \end{tikzpicture}
    }}
    \caption{Breakdown of MIMOSA for the depth-2  \texttt{either} version of Chopsticks, colored by ground-truth assignments. MIMOSA learns an initial 4D softplus AE representation (left), decomposes it into lower-dimensional components by grouping together neighboring points with similar local SVDs (second from left), merges them over longer distances while discarding outliers (second from right), and finally uses enclosure relationships to infer a hierarchy (right). In this case, correspondence between the assignment of points to learned components vs. ground-truth is very close (99.8\% purity, covering 93.7\% of the training set, and with no $H$-error---see \S\ref{sec:mimosa-metrics} for definitions of these metrics). %Initial representations tended to place points with the same assignments on smooth but intersecting manifolds, with higher local density at lower dimensions. MIMOSA was able to disentangle these components and estimate the correct hierarchical relationships.
    Similar examples are shown for other datasets in Figs.~\ref{fig:spaceshapes-manifolds}-\ref{fig:chopsticks322-manifolds} of the Appendix.}
    \label{fig:chopsticks-manifolds}
\end{figure*}

The goal of our first algorithm, MIMOSA (\textbf{M}ulti-manifold \textbf{I}so\textbf{M}ap \textbf{O}n \textbf{S}mooth \textbf{A}utoencoder), is to learn a hierarchy $\hat{H}$ from data, as well as an assignment vector $\hat{A}_n$ of data points to hierarchy leaves.  
%This algorithm is topological in nature; once MIMOSA identifies the structure within the latent representation, COFHAE (Section~FILL) learns the parameters. 
MIMOSA consists of the following steps (see Appendix for Algorithms \ref{algo:local-svd}-\ref{algo:construct-hier} and complexity, and Fig.~\ref{fig:chopsticks-manifolds} for a detailed example): 

\textbf{Dimensionality Reduction (Algorithm~\ref{algo:mimosa}, line 1):} We start by performing an initial reduction of $X$ to $Z$ using a flat autoencoder. While we could start with $Z=X$, performing this reduction saves computation as later steps (e.g. finding neighbors) scale linearly with $|Z|$. Although this requires choosing $|Z|$, we find the exact value is not critical as long as it exceeds the (max) intrinsic dimensionality of the data. We also find it important to use differentiable activation functions (e.g. Softplus rather than ReLU) to keep latent manifolds smooth; see Fig.~\ref{fig:relu-vs-softplus} for more.
    
\textbf{Manifold Decomposition (Algorithms~\ref{algo:local-svd}-\ref{algo:merge-components}):} We next decompose $Z$ into a set of manifold ``components'' by computing SVDs locally around each point and merging neighboring points with sufficiently similar subspaces. 
We then perform a second merging step over longer lengthscales, combining equal-dimensional components with similar local SVDs along their nearest boundary points and discarding small outliers, which we found was necessary to handle interstitial gaps when two manifolds intersect.
The core of this step is based on a multi-manifold learning method~\citep{mahapatra2017s}, but we make efficiency as well as robustness improvements by combining ideas from RANSAC~\citep{fischler1981random} and contagion dynamics~\citep{mahler2020contagion}. The merging step is a new contribution.

It bears emphasis that manifold decomposition, which groups points based on the similarity of local principal components, is distinct from clustering, which groups points based on proximity. In the examples we consider, even hierarchical iterative clustering methods like OPTICS \citep{ankerst1999optics} will not suffice, as nearby points may lie on different manifolds.

\textbf{Hierarchy Identification (Algorithm~\ref{algo:construct-hier}):} Finally, we construct a tree by drawing edges from low-dimensional components to the higher-dimensional components that best ``enclose'' them, which we define using a ratio of inter-component to intra-component nearest neighbor distances; we believe this is novel. We use this tree and the component dimensionalities to construct a dimension hierarchy and a set of assignments from points to paths, which we output.

\textbf{Hyperparameters:} Each of these steps has several hyperparameters, and we provide a full listing and sensitivity study in \S\ref{sec:mimosa-hypers}. 
The one we found most critical was the minimum SVD similarity to merge neighboring points.% (the $\epsilon$ from \citet{mahapatra2017s}). 

\subsection{Training Autoencoders with COFHAE}

\begin{algorithm}
    \caption{COFHAE($X$)}
    \label{algo:cofhae}
\begin{algorithmic}[1]
    \STATE hierarchy, assignments = MIMOSA($X$)\, \textcolor{gray}{\# Algorithm~\ref{algo:mimosa}}
    \STATE $\mathrm{HAE}_{\theta} = \text{init\_hierarchical\_autoencoder}(\mathrm{hierarchy})$
    \STATE $D_{\psi} = \text{init\_discriminator}()$
    \FOR{$x, a \sim \mathrm{minibatch}(X, \mathrm{assignments})$}
    \STATE $a', z = \mathrm{HAE}_{\theta}.\mathrm{encode(}x; \tau)$\, \textcolor{gray}{\text{\# Algorithm~\ref{algo:encode}}}
    \STATE $x' = \mathrm{HAE}_{\theta}.\mathrm{decode}(\mathrm{concat(}a', z\mathrm{))}$\, \textcolor{gray}{\text{\# normal NN}}
    \STATE $z' = \mathrm{copy}(z)$
        
    \FOR{$i = 1..|z_0|$}
    \STATE $\mathrm{shuffle}\, z'_{:,i}\, \mathrm{over\,minibatch \,indices\,where}\, \mathtt{on}(z_{:,i})$
    \ENDFOR
    \STATE $\mathcal{L}_{\theta} = \mathcal{L}_x(x',x) + \lambda_1 \mathcal{L}_a(a',a)  - \lambda_2 \log \frac{D_{\psi}(z)}{1-D_{\psi}(z)}$
    \STATE $\mathcal{L}_{\psi} = {-}\log D_{\psi}(z') - \log(1-D_{\psi}(z))$
    \STATE $\theta = \mathrm{descent\_step}(\theta, \mathcal{L}_{\theta})$
    \STATE $\psi = \mathrm{descent\_step}(\psi, \mathcal{L}_{\psi})$
    \ENDFOR
    \STATE \textbf{return} $\mathrm{HAE}_{\theta}$
\end{algorithmic}
\end{algorithm}

Our first stage, MIMOSA, gives us the hierarchy and assignments of data down it.  In the second stage, COFHAE (\textbf{CO}nditionally \textbf{F}actorized \textbf{H}ierarchical \textbf{A}uto\textbf{E}ncoder, Algorithms~\ref{algo:cofhae} and \ref{algo:encode}), we learn an autoencoder that respects this hierarchy via (differentiable) masking operations that impose structure on flat representations.  

\textbf{Hierarchical Encoding (Algorithm~\ref{algo:encode}):} Instances $x$ pass through a neural network encoder to an initial vector $z_{pre}$, whose dimensions correspond to both continuous and categorical dimensions. We then pass each set of categoricals through a softmax with temperature $\tau$, and use them to recursively mask the entirety of $z_{pre}$ based on the hierarchy. We finally split this masked vector into a continuous vector $z$ and a list of estimated assignments $a'$, outputting both.

\textbf{Supervising Assignments:} Although we lack ground-truth during training, we do have assignments $a$ from MIMOSA (for at least a subset of the dataset). We add a penalty $\mathcal{L}_a(a', a)$, weighted by $\lambda_1$, to make encoded $a'$ match $a$.

\textbf{Conditional Factorization:} 
 \citet{kim2018disentangling} penalize the total correlation (TC) between dimensions of flat continuous representations $z$ with two tricks. First, noting that TC is the KL divergence between $q(z)$ (the joint distribution of the encoded $z$) and $\bar{q}(z) \equiv \prod_{j=1}^{|z|} q(z_j)$ (the product of its marginals), they approximate samples from $\bar{q}(z)$ by randomly permuting the values of each $z_i$ across batches~\citep{arcones1992bootstrap}. Second, they approximate the KL divergence between the two distributions using the density ratio trick~\citep{sugiyama2012density} on an auxiliary discriminator $D_\psi(z)$, where $KL(q(z)||\bar{q}(z)) \approx \log \frac{D_{\psi}(z)}{1-D_{\psi}(z)}$ if $D_\psi(z)$ outputs accurate probabilities of $z$ having been sampled from $\bar{q}$.
% FDV: Can you describe the key innovation here more succinctly/conceptually without the notation, and then version with more notation goes in appendix?
% ASR: I think it's important to keep it so that D_\psi is defined in the algorithm box!
We adopt a similar approach, except instead of permuting each $z_i$ across the full batch $\mathcal{B}$, we only permute it where it is \emph{active}, i.e. $\mathcal{B} \cap \mathtt{on}(z_i)$ (defined using the hardened version of the mask). This approximates a hierarchical version of $\bar{q}(z)$ where each dimension distribution is a mixture of 0 and the product of its \emph{active} marginals. $D_\psi(z)$ then lets us estimate the KL between this distribution and $q(z)$, which we penalize and weight with $\lambda_2$.

This approach presumes ground-truth continuous variables should be conditionally independent given categorical values, which is a major assumption. However, it is less strict than the assumption taken by many disentanglement methods, i.e. that continuous variables are independent marginally, and it may remain useful as an inductive bias.%, e.g. for causal graphs.

\section{Hierarchical Disentanglement Metrics}\label{sec:metrics}

In this section, we develop metrics for quantifying how well learned representations and hierarchies match ground-truth.

\subsection{Desiderata and Invariances}\label{sec:metric-desiderata}
Our goal in designing metrics is to measure whether we have learned the ``right representation,'' both in terms of global structure and specific variable correspondences.
In an ideal world, we would measure whether a learned representation $Z$ is identically equal to a ground-truth $V$.
However, most existing disentanglement metrics are invariant to permutations, so that dimensions $V_i$ can be reordered to match different $Z_j$, as well as univariate transformations, so that the values of $Z_j$ do not need to be identical to $V_i$. In the case of methods like the SAP score~\citep{kumar2017variational}, these univariate transformations must be linear, but as the uniformity of scaling can be arbitrary, we permit general nonlinear transformations, as long as they are 1:1, or invertible.

Hierarchical representations have an additional ambiguity about the right ``vertical'' placement of continuous variables. For example, on Spaceshapes, the $\mathtt{phase}$, $\mathtt{shine}$, and $\mathtt{angle}$ variables could all be ``merged up'' to a single top-level variable whose effect changes based on $\mathtt{shape}$. Alternatively, $\mathtt{x}$ and $\mathtt{y}$ position could be ``pushed down'' and duplicated for each shape despite their analogous effects (see Fig.~\ref{fig:spaceshapes-alts} for an illustration).
In terms of our user interface analogy from Fig.~\ref{fig:hier-reps-html} (or our specific \href{https://hreps.s3.amazonaws.com/viz/index.html?dataset=spaceshapes&model=supervised}{implementation}), ``merge up'' and ``push down'' transformations correspond to moving sliders into or out of outlined groups, but keeping their effects on the outputs the same, as well as preserving the structure of nested radio buttons. To a user interacting with such representations, they would appear almost identical, except some slider labels might change with radio button settings.
Because of this functional near-equivalence,
we defer the problem of deciding the most natural vertical placement of continuous variables to future work, and make our main metrics invariant to them.

\subsection{MIMOSA Metrics: $H$-error, Purity, Coverage}
\label{sec:mimosa-metrics}
 
The first metric we use to evaluate MIMOSA is \textbf{$H$-error}, which measures whether learned hierarchy $\hat{H}$ has the same essential structure as the ground-truth hierarchy $H$. We define $H$-error in terms of the tree edit distance of \citet{zhang1989simple} (i.e. minimum number of insertions, edits, or deletions to transform $H$ into $\hat{H}$), but between normalized ``merged up'' representations of each hierarchy; details are in \S\ref{sec:h-error-details}.
This metric is 0 if and only if both hierarchies are the same up to the transformations described in \S\ref{sec:metric-desiderata}.

The second MIMOSA metric is \textbf{purity}, which measures whether the assignments output by MIMOSA match ground-truth. To compute it, we iterate over each leaf in $\hat{H}$, find the leaf in $H$ to which most of its assigned points belong, and then compute the fraction that belong to the majority. Then we average these scores across $\hat{H}$, weighting by the number of points in each leaf. This metric only falls below 1 if leaves contain points with different ground-truth assignments.

The final metric we use to evaluate MIMOSA is \textbf{coverage}. Since MIMOSA discards small outlier components, it is possible that the final set of assignments will not cover the full training set. If almost all points are discarded this way, the other metrics may not be meaningful. As such, we measure coverage as the fraction of the training set which is not discarded. We note that hyperparameters can be tuned to ensure high coverage without knowing ground-truth assignments.

\subsection{COFHAE Metrics: $R^4$ and $R^4_c$ Scores }\label{sec:cofhae-metrics}

Per our desiderata, we seek to check whether every ground-truth variable $V_i$ can be mapped invertibly to some learned dimension $Z_j$. 
As a preliminary definition, we say that a learned $Z_j$ \emph{corresponds} to a ground-truth $V_i$ over some set $\mathcal{S} \subseteq \mathbb{R}$ if a bijection between them exists; that is, \begin{equation}\label{eqn:disent}
\begin{split}
    \exists\, f(\cdot): \mathcal{S} \to \mathbb{R} \text{ s.t. }
    f(V_i) &= Z_j
   \text{ and } f^{-1}(Z_j) = V_i
\end{split}
\end{equation}We say that $Z$ \emph{disentangles} $V$ if all $V_i$ have a corresponding $Z_j$.
To measure the extent to which bijections exist, we can simply try to learn them (over random splits of many paired samples of $V_i$ and $Z_j$). Concretely, for each pair of learned and true dimensions, we train univariate models to map in both directions, compute their coefficients of determination ($R^2$), and take their geometric mean: 
\begin{equation}
    \begin{split}
       f &\equiv \min_{f \in \mathcal{F}} \mathbb{E}_{\mathrm{train}}\left[ (f(X)-Y)^2 \right] \\
         R^2(X{\to}Y) &\equiv \mathbb{E}_{\mathrm{test}} \left[
        1-\frac{\sum (f(X)-Y)^2 }{\sum (\mathbb{E}[Y]-Y)^2} \right]\\
        R^2(X{\leftrightarrow}Y) &\equiv \sqrt{\lfloor R^2(X{\to}Y) \rfloor_{+} \lfloor R^2(Y{\to}X) \rfloor_{+}},\\
\end{split}
\end{equation}where we average over train/test splits (we use 5), assume $\mathcal{F}$ is sufficiently flexible to contain the optimal bijection (we use gradient-boosted decision trees), and assume our dataset is large enough to reliably identify $f \in \mathcal{F}$.
In the limit, $R^2(X{\leftrightarrow}Y)$ can only be 1 if a bijection exists, as any region of non-zero mass in the joint distribution of $X$ and $Y$ where this is false would imply $\mathbb{E}[(f(X)-Y)^2] >0$ or $\mathbb{E}[(f(Y)-X)^2] >0$. 
In the special case that $Y$ is discrete rather than continuous, we use classifiers for $f$ and accuracy instead of $R^2$, but the same argument holds. 
%Note that we also considered using rank correlation and normalized mutual information~\citep{chen2018isolating}, but the latter is technically undefined for continuous variables, while the former fails when checking if continuous variables approximately correspond to (arbitrarily-ordered) categoricals.

To measure whether a \emph{set} of variables $Z$ disentangles another \emph{set} of variables $V$, we check if, for each $V_i$, there is at least one $Z_j$ for which $R^2(V_i \leftrightarrow Z_j)=1$: \begin{equation}
     R^4(V, Z) \equiv \frac{1}{|V|} \sum_{i} \max_j R^2(V_i \leftrightarrow Z_j),
\end{equation} We call this the ``right-representation'' $R^2$, or $R^4$ score. Note that this metric is related to the existing SAP score~\citep{kumar2017variational}, except we allow for nonlinearity, require high $R^2$ in both directions, and take the maximum over each score column rather than the difference between the top two entries (which avoids assuming ground-truth is factorized).

Although $R^4$ is useful for measuring correspondence between sets of variables that are both always active, it does not immediately apply to hierarchical representations unless inactive variables are represented somehow, e.g. as 0 (an arbitrary implementation decision that affects $R^2$ by changing $\mathbb{E}[Y]$). It also lacks invariance to merge-up and push-down operations.
Instead, we seek \emph{conditional correspondence} between $V_i$ and a set of dimensions in $Z$, defined as \begin{equation}
    \begin{split}
        \text{for}&\,\text{all}\, V_i\in\mathtt{on}(V_i)\, \exists\, \mathcal{Z}_{i} = \{Z_j, Z_k,\hdots\}\, \mathrm{s.t.} \\
         &(a)\, V_i\,\text{corresponds to}\,Z_j\,\text{over}\,\mathtt{on}(V_i) \cap \mathtt{on}(Z_j),\\
        &(b)\, \mathtt{on}(Z_j) \cap \mathtt{on}(Z_k) = \emptyset\,\text{for all}\,j\neq k, \text{and}\\
        &(c)\, \textstyle\bigcup_{z\in \mathcal{Z}_{i}} \mathtt{on}(z) = \mathtt{on}(V_i),
    \end{split}
\end{equation}
or rather that we can find some tiling of $\mathtt{on}(V_i)$ into regions where it corresponds 1:1 with different $Z_j$ which are never active simultaneously. This allows for one $Z_j$ to correspond to non-overlapping elements of $V$ (e.g. merging up), as well as for one $V_i$ to be modeled by non-overlapping elements of $Z$ (e.g. pushing down).

We can then formulate a conditional $R_c^4$ score which quantifies how closely conditional correspondence holds: 
\begin{equation*}
\begin{split}
R^2_c(V_i, g)
\equiv& \max\bigg(
\max_{j\in g}\Big(
    R^2\big(V_i {\leftrightarrow} Z_j \big| \mathtt{on}(V_i) \cap \mathtt{on}(g)\big),\\
    &\displaystyle\sum_{g' \in \mathtt{children}(Z_j)}R^2_c(V_i, g') \frac{\left| \mathtt{on}(V_i) \cap \mathtt{on}(g') \right|}{|\mathtt{on}(V_i)|}  
\Big)\bigg),
\end{split}
\end{equation*}
for a dimension group $g$; the overall disentanglement is:
\begin{equation}
    R^4_c(V{\leftrightarrow}Z) \equiv \frac{1}{|V|}\sum_{i=1}^{|V|} R^2_c(V_i, \mathtt{root}(Z)).
\end{equation}
In the special case that $V$ and $Z$ are flat, $R^4_c$ reduces to $R^4$.
We note that even for flat representations, the $R^4$ score may be a useful measure of disentanglement when ground-truth variables are not factorized.

\section{Experimental Setup}

\textbf{Benchmarks:} We ran experiments on nine benchmark datasets: Spaceshapes, and eight variants of Chopsticks (varying \texttt{slope}, \texttt{intercept}, \texttt{both}, and \texttt{either} at recursion depths of 2 and 3). See \S\ref{sec:benchmarks} for more details, and Fig.~\ref{fig:mimosa-noise} for preliminary experiments on noisy data.

\begin{figure}[h!]
    \centering
    \includegraphics[width=0.95\linewidth]{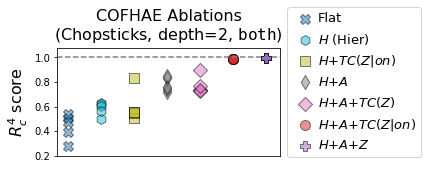}
    \caption{Ablation study for COFHAE on the depth-2 $\mathtt{both}$ version of Chopsticks (over 5 restarts). Hierarchical disentanglement is low for flat AEs (Flat); adding the ground-truth hierarchy $H$ improves it (Hier $H$), as does also adding supervision for ground-truth assignments $A$ ($H{+}A$). Adding a FactorVAE-style marginal TC penalty ($H{+}A{+}TC(Z)$) does not appear to help disentanglement, but making that TC penalty conditional ($H{+}A{+}TC(Z|on)$, i.e. COFHAE) brings it close to the near-optimal disentanglement of a hierarchical model whose latent representation is fully supervised ($H{+}A{+}Z$). However, the hierarchical conditional TC penalty fails to produce this same disentanglement without any supervision over assignments ($H{+}TC(Z)$).}
    \label{fig:cofhae-ablations}
\end{figure}

\textbf{Algorithms:} In addition to COFHAE with MIMOSA, we trained a number of flat baselines. As fully continuous baselines, we trained autoencoders (AE), variational autoencoders \citep{kingma2013auto}  (VAE), the $\beta$-total correlation autoencoder \citep{chen2018isolating} (TCVAE), and FactorVAE \citep{kim2018disentangling}. As mixed discrete-continuous baselines, we trained JointVAE \citep{dupont2018learning} and CascadeVAE \citep{jeong2019learning}, providing them with the ground-truth structure of discrete variables.\footnote{Note that CascadeVAE only supports a single categorical variable, but we give it cardinality equal to the total number of paths down the true hierarchy.} Finally, we ran COFHAE ablations using the ground-truth hierarchy and assignments, testing all possible combinations of loss terms and comparing conditional vs. marginal TC penalties; results are in Fig.~\ref{fig:cofhae-ablations}.
See \S\ref{sec:training-details} for training and model details.
%Architecturally, we used the dSprites convolutional architecture from \citet{burgess2018understanding} with Bernoulli loss on Spaceshapes and 256$\times$256 fully-connected layers with Gaussian loss on Chopsticks. See Appendix~\ref{sec:training-details} for training details.

\begin{figure}[h!]
    \centering
    \includegraphics[width=\linewidth]{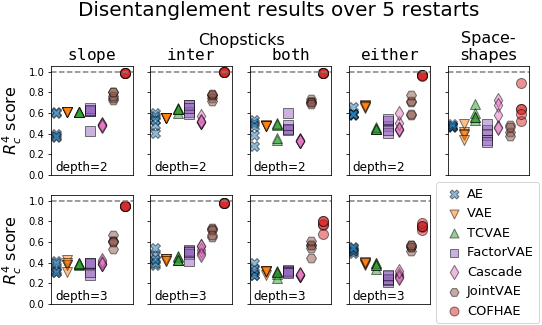}
    \caption{Hierarchical disentanglement results for representation learning methods (baselines and COFHAE + MIMOSA) over all nine datasets. COFHAE almost perfectly disentangles ground-truth on the six simplest versions of Chopsticks, with some degradations on the two most complex versions (with very deep hierarches) and on Spaceshapes (with a shallower hierarchy, but higher-dimensional inputs). Baseline methods were generally much more entangled, though JointVAE, $\beta$-TCVAE, and CascadeVAE are competitive in certain cases.}
    \label{fig:disent-all}
\end{figure}

\begin{figure}
    \centering
    \includegraphics[width=\linewidth]{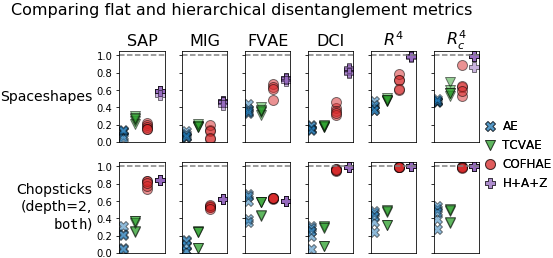}
    \caption{Comparison of disentanglement metrics across two datasets and four models. Only $R^4$ and $R^4_c$ correctly and consistently award near-optimal scores to the supervised H+A+Z model.}
    \label{fig:disent-metric-comp}
\end{figure}

\textbf{Metrics:} To evaluate hierarchies, we computed purity, coverage, and $H$-error (\S\ref{sec:mimosa-metrics}). Results are in Table~\ref{tab:mimosa-results}. To measure disentanglement, we primarily use $R^4_c$ (\S\ref{sec:cofhae-metrics}); results for all datasets and models are in Fig.~\ref{fig:disent-all}. We also compute the following baseline metrics: the SAP score \citep{kumar2017variational} (SAP), the mutual information gap \citep{chen2018isolating} (MIG, estimated using 2D histograms), the FactorVAE score \citep{kim2018disentangling} (FVAE), and the DCI disentanglement score \citep{eastwood2018framework} (DCI). Most implementations were adapted from \texttt{disentanglement\_lib} \citep{locatello2018challenging}. We also compute our marginal $R^4$ score. Results across metrics are shown for a subset of datasets and models in Fig.~\ref{fig:disent-metric-comp}.

\textbf{Hyperparameters:} COFHAE is only given instances $X$, which complicates cross-validation. However, we can still tune its hyperparameters to ensure assignments $a'$ match MIMOSA outputs $a$ and reconstruction loss for $x$ is low (which fail to can happen if the adversarial term dominates). Over a grid of $\tau$ in $\{\frac{1}{2}, \frac{2}{3}, 1\}$, $\lambda_1$ in $\{10,100,1000\}$, and $\lambda_2$ in $\{1,10,100\}$, we select the model with the lowest \emph{training} reconstruction loss $\mathcal{L}_x$ from the $\frac{1}{3}$ with the lowest assignment loss $\mathcal{L}_a$.
For MIMOSA, hyperparameters can be tuned to ensure high coverage (purity and $H$-error require side-information); see \S\ref{sec:mimosa-hypers} for more details. 

For our baselines, we show results at $\beta{=}5$ for TCVAE, $\gamma{=}10$ for FactorVAE, $\beta{=}1, C_z{=}C_c{=}10$ for JointVAE, and $\beta_\ell{=}2$ for CascadeVAE (with other hyperparameters set to the same settings as the original paper). However, we tested each method across a variety of strength and capacity hyperparameters, and show more complete results in Fig.~\ref{fig:baseline-hypers}.

%; architectural and training details are in Appendix~\ref{sec:training-details}. %Code and data will be public.

%As baseline algorithms, we consider 
%These baselines are meant to be illustrative of existing methods but not exhaustive; to provide additional comparison, we also run many possible ablations of COFHAE (see Figure~\ref{fig:cofhae-ablations}).

\begin{table*}[h!]
    \centering
    \begin{tabular}{|c||c|c|c|c||c|c|c|c||c|}\hline
   MIMOSA  & \multicolumn{4}{c||}{Chopsticks, depth=2} &  \multicolumn{4}{c||}{Chopsticks, depth=3} & Space- \\\cline{2-9}
   Metric & \texttt{inter} & \texttt{slope} & \texttt{both} & \texttt{either} & \texttt{inter} & \texttt{slope} & \texttt{both} & \texttt{either} & shapes \\\hline\hline
%   Purity & $1.0{\pm}0.0$ & $1.0{\pm}0.0$ & $1.0{\pm}0.0$ & $1.0{\pm}0.0$ & $.98{\pm}0.0$ & $.95{\pm}0.0$ & $.94{\pm}0.0$ & $.93{\pm}0.0$ & $1.0{\pm}0.0$ \\\hline
% Coverage & $.99{\pm}0.0$ & $.99{\pm}0.0$ & $.96{\pm}0.0$ & $.93{\pm}0.0$ & $.98{\pm}0.0$ & $.98{\pm}0.0$ & $.82{\pm}0.01$ & $.75{\pm}0.01$ & $1.0{\pm}0.0$ \\\hline
%$H$-error & $0.0{\pm}0.0$ & $0.0{\pm}0.0$ & $0.0{\pm}0.0$ & $0.0{\pm}0.0$ & $0.0{\pm}0.0$ & $0.0{\pm}0.0$ & $0.0{\pm}0.0$ & $2.6{\pm}1.34$ & $0.0{\pm}0.0$ \\\hline
Purity & $1.0{\pm}0.0$ & $1.0{\pm}0.0$ & $1.0{\pm}0.0$ & $1.0{\pm}0.0$ & $.98{\pm}0.0$ & $.95{\pm}0.0$ & $.94{\pm}0.0$ & $.93{\pm}0.0$ & $1.0{\pm}0.0$ \\\hline
 Coverage & $.99{\pm}0.0$ & $.99{\pm}0.0$ & $.96{\pm}0.0$ & $.94{\pm}0.0$ & $.98{\pm}0.0$ & $.98{\pm}0.0$ & $.82{\pm}0.01$ & $.75{\pm}0.01$ & $1.0{\pm}0.0$ \\\hline
$H$-error & $0.0{\pm}0.0$ & $0.0{\pm}0.0$ & $0.0{\pm}0.0$ & $0.0{\pm}0.0$ & $0.0{\pm}0.0$ & $0.0{\pm}0.0$ & $0.0{\pm}0.0$ & $2.40{\pm}0.89$ & $0.0{\pm}0.0$ \\\hline
    \end{tabular}
    \caption{MIMOSA results across all datasets, with means and standard deviations across 5 restarts. In general, MIMOSA components contained points only from single ground-truth sets of paths (purity), were inclusive of most points in the training set (coverage), and resulting in perfectly accurate hierarchies ($H$ errors), with the greatest or only exception being the Chopsticks depth-3 \texttt{either} dataset (where we tended to miss 2-3 of the 8 deepest 3D components).}
    \label{tab:mimosa-results}
\end{table*}

\section{Results and Discussion}

\textbf{MIMOSA consistently recovered the right hierarchies.}
Per Table~\ref{tab:mimosa-results}, we consistently found the right hierarchy for all datasets except depth-3 $\mathtt{either}$-Chopsticks, but even there results were close, generally recovering 12 out of 14 nodes (see Fig.~\ref{fig:chopsticks322-manifolds} for more details). Purity and coverage were also high, often near perfect as in Spaceshapes or depth-2 Chopsticks.

\textbf{COFHAE significantly outperformed baselines.}
Per Fig.~\ref{fig:disent-all}, COFHAE $R^4_c$ scores were near-perfect for 6 out of 9 datasets, and were highest on all (both in terms of mean and maximum). On Spaceshapes and the depth-3 $\mathtt{either}$ and $\mathtt{both}$ versions of Chopsticks, scores were slightly worse. Part of this suboptimality could be due to non-identifiability. For Spaceshapes and the $\mathtt{both}$ versions of Chopsticks, dimension group nodes contain multiple continuous variables, which even conditionally can be modeled by multiple factorized distributions \citep{locatello2018challenging}. However, optimization issues could also be at fault, as we do not see suboptimal $R^4_c$ on Chopsticks until a depth of 3, and even supervised $H{+}A{+}Z$ models occasionally fail to converge on Spaceshapes. \citet{kim2018disentangling} note that the relatively low-dimensional discriminator used by FactorVAE is easier to optimize than the generally high-dimensional discriminators used in GANs, which are notoriously tricky to train \citep{mescheder2018training}. In our case, flattened hierarchy vectors can be high-dimensional (e.g. Fig.~\ref{fig:chopsticks-322-qual}), and in any given batch, instances corresponding to different paths down the hierarchy may have different numbers of samples (potentially requiring larger batch sizes or stratified sampling to ensure sufficient coverage). Finally, alongside non-identifiability and optimization issues, MIMOSA errors (e.g. merge-up/push-down differences for Spaceshapes and suboptimal purity and coverage for Chopsticks) also may play a role, as evidenced by performance improvements in our full COFHAE ablations in Fig.~\ref{fig:ablations-all}. 
Despite all of these issues, COFHAE is still closer to optimal, at best and on average, than any of our baseline algorithms (even on Spaceshapes, where it is possible for a flat representation to disentangle all features except jet length). We note also that our baselines often performed \emph{worse} with increasing 
disentanglement penalty strength (Fig.~\ref{fig:baseline-hypers}), with the closest COFHAE competitor, JointVAE, achieving its best results at its minimal tested value $\gamma{=}1$ (i.e. equivalent to a normal VAE).
These results are consistent with the fact that minimizing marginal rather than conditional TC on these datasets \emph{prevents} models from learning the right representation.
%$R^4_c$ does not increase all the way to 1 on these three hardest datasets, suggesting non-identifiability and/or adversarial optimization issues are still at play. However, they are still much closer to optimal than any of our baselines.

\textbf{$R^4_c$ provides more insight into disentanglement than baselines.}
One way to evaluate an evaluation metric is to test it against a precisely known quantity. In this case, we know the $H{+}A{+}Z$ model, whose encoder is supervised to match ground-truth, should receive a near-perfect score. The only metrics to do this consistently are $R^4$ and $R^4_c$. Note that the DCI disentanglement score, based on the entropy of normalized feature importances from an estimator predicting single ground-truth factors from all learned dimensions, comes close. Intuitively, this metric could behave similarly to $R^4$ if its estimator was trained to be sparse (placing importance on as few dimensions as possible). However, using $R^2$s of univariate estimators is more direct, and also incorporates information from the DCI informativeness score.

Another way to evaluate an evaluation metric is to test whether quantitative differences capture salient qualitative differences. To this point, specifically to compare $R^4$ and $R^4_c$, we consider several examples in Fig.~\ref{fig:cofhae-spaceshapes-qualitative} and Fig.~\ref{fig:hier-latent-traversal}. First, we see that for the Spaceshapes COFHAE model in Fig.~\ref{fig:cofhae-spaceshapes-qual} (or \href{https://hreps.s3.amazonaws.com/viz/index.html?dataset=spaceshapes&model=cofhae}{here}), its $R^4_c$ score (0.89) is higher than its $R^4$ (0.79). This increase is due to the fact that $R^4$ penalizes ``push-down'' differences (\S\ref{sec:metric-desiderata}) between the learned and true factors representing $\mathtt{x}$ and $\mathtt{y}$ position, while $R^4_c$ is invariant to them. However, the overall increase is less dramatic than one might expect due to modest decreases in correspondence scores for other dimensions (e.g. $0.98{\to}0.89$ for $\mathtt{jetlen}$), which occur because $R^4_c$ is not biased by spurious equality between dimensions which are both inactive.
%These decreases are due to the fact that $R^4$ scores for each $V_i$ are computed over the full dataset rather than subsets where $V_i$ and candidate $Z_j$s are active. Outside these subsets, inactive values are implemented as 0 (despite being technically undefined), and when such $V_i$ and $Z_j$ are both inactive (which in some cases occurs $2/3$ or $5/6$ of the time), they are treated as equal. This incidental equality spuriously increases $R^4$ but not $R^4_c$. For example, in the case of $\mathtt{jetlen}$, which has a close but not completely 1:1 relationship to $Z_{12}$, its contribution to $R^4_c$ is 0.89 (which seems consistent with the histograms) while its contribution to $R^4$ is 0.98 (which seems too high).
Another example of a difference between $R^4$ and $R^4_c$ (illustrating invariance to ``merging up'' rather than ``pushing down'') is for the Spaceshapes $\beta$-TCVAE in Fig.~\ref{fig:tcvae-spaceshapes-qual}. In this case, histograms show that one $\beta$-TCVAE variable ($Z_3$) corresponds closely to both moon $\mathtt{phase}$ and star $\mathtt{shine}$ (and to a lesser extent, $\mathtt{jetlen}$), only one of which is active at a time. The $R^4$ score (0.47) assigns low scores to these correspondences, but $R^4_c$ (0.69) properly factors them in.

\textbf{COFHAE and MIMOSA subcomponents improve performance.}
%Although the methods we introduce contain many moving parts, results in Fig.~\ref{fig:cofhae-ablations} and Fig.~\ref{fig:ablations-all} suggest all COFHAE components count, and in particular that conditional factorization penalties produce more disentangled solutions than marginal ones.
Though COFHAE contains many moving parts, results in Fig.~\ref{fig:cofhae-ablations} and Fig.~\ref{fig:ablations-all} suggest they all count. Autoencoders only achieve optimal disentanglement if provided with the hierarchy, assignments, and a conditional (not marginal) penalty on the TC of continuous variables; no partial subset suffices.
In the Appendix, Fig.~\ref{fig:mimosa-ablations} shows ablations and sensitivity analyses for MIMOSA that validate its subcomponents are important as well.

\subsection{Remark on Identifiability and Parsimony}\label{sec:mdl}

From \citet{locatello2018challenging}, we know all forms of TC minimization permit multiple solutions (though they often improve disentanglement empirically, especially when ground-truth factors are non-Gaussian). However, what about the other components of our method, such as MIMOSA?

MIMOSA does not minimize an objective function, so questions of identifiability might seem moot. However, we could reformulate it as trying to find a \emph{small} set of \emph{low-dimensional} and \emph{bounded-curvature} manifolds that \emph{approximately contain} a \emph{large fraction} of the data. More concretely, we could place penalties or constraints on, e.g., the cardinality, dimensionality, and mean or percentiles of error and principal curvature magnitudes over the set. Such a problem might well be identifiable (up to the transformations discussed in \S\ref{sec:metric-desiderata}), though analyzing it is beyond the scope of this work.

However, perhaps a better-motivated formulation that covers both MIMOSA and COFHAE would be to return to minimum description length (MDL)---the same problem that motivated much of the initial research into factorized representations \citep{barlow1961possible,zemel1994minimum}.
As an example, assume we are given a dataset of $N$ instances, $\frac{7}{8}$ of which lie on a 1D manifold, and $\frac{1}{8}$ of which lie on an 8D manifold. If we must encode instances as flat vectors of 32-bit floats, those vectors will need to be at least 8D for accurate reconstruction, meaning the dataset's description length will be $8 * 32 * N = 256N$ bits (plus the size of the model, which is negligible for sufficiently large $N$). However, if we use a disentangled hierarchical representation, we need either 1 or 8 floats to represent each instance (plus a single bit to distinguish between them). In that case, the description length would be $(\frac{1}{8} * 8 * 32) + (\frac{7}{8} * 1 * 32) + 1)N = 61N$ bits, which is minimal (assuming the model is not much larger). 
The problem of learning factorized representations \emph{within} each manifold might remain non-identifiable, but the MDL argument for doing so remains the same as in \citet{zemel1994minimum}.
This example suggests that (disentangled) hierarchical representations might spontaneously emerge as the (partially identifiable) solution to MDL objectives, at least for datasets that lie on multiple manifolds.
%There are many other limitations with our methods, especially in the presence of noise (Fig.~\ref{fig:mimosa-noise}), 

%\paragraph{Additional analysis:} In the Appendix, we dig deeper into the dimension-by-dimension correspondences between different models and ground-truth (Fig.~\ref{fig:cofhae-spaceshapes-qualitative} and Fig.~\ref{fig:chopsticks-322-qual}) and MIMOSA manifolds (Figs.~\ref{fig:spaceshapes-manifolds}, \ref{fig:chopsticks201-manifolds}, and \ref{fig:chopsticks322-manifolds}). We also explore ReLU vs. Softplus activations (Fig.~\ref{fig:relu-vs-softplus}) and flat vs. hierarchical disentanglement metrics (Fig.~\ref{fig:disent-score-comp}).

\section{Conclusion}

In this work, we introduced a novel formulation of hierarchical disentanglement, where ground-truth representation dimensions are organized into a tree and activated or deactivated based on the values of categorical dimensions. We presented benchmarks, algorithms, and metrics for learning and evaluating such hierarchical representations. 
%Although our algorithms perform well on most of our benchmarks, there is still room for improvement, and see our contribution more as opening up a problem than closing in on a solution.

% 1. Improving the existing method:
%% handle noise
%% theoretical results
% 2. Extending it:
%% DAGs
%% Object reps
% 3. Real world

There are a number of promising avenues for future work.
One is extending our methods to handle a wider variety of underlying structures, e.g. dimension DAGs, or integrating our methods with object representation techniques to better model generative processes involving ordinal variables or unordered sets~\citep{locatello2020object}.
Another is to better solve or understand hierarchical disentanglement as we have already formulated it, e.g. by improving robustness to noise (Fig.~\ref{fig:mimosa-noise}) or providing a better theoretical understanding of identifiability, perhaps through the lens of description length.
Finally, there are ample opportunities to apply these techniques to real-world data that we expect to have hierarchical multiple-manifold structure, such as patient phenotype or population genetics datasets.

More generally, we feel it is important for representation learning to move beyond flat vectors, and work towards explicitly modeling the rich structure contained in the real world.
Symbolic AI and cognitive science researchers have made compelling arguments that future AI progress should be evaluated not by improvements in accuracy or reconstruction error, but by how well models build their own interpretable models of the world~\citep{lake2017building}. Our work takes steps in this direction.
%In this work, we provide a novel framework for both evaluating and learning such models.
%The novel benchmarks, algorithms, and metrics we present in this paper may bring us closer to that goal.

%Such a shift would help address criticisms from the cognitive science and symbolic AI communities that machine learning focuses too much on pattern recognition at the expense of 

%While flat vectors are convenient for computation and analysis, 

%While flat vectors may be convenient for computation or analysis,
%The novel results we present suggest that adding structure to representations can improve interpretability, reduce non-identifiability,
%and potentially help bridge longstanding gaps between machine learning and symbolic AI.
%While flat vectors may be convenient for computation or analysis, 

%In a more general sense, we feel it is important for representation learning and machine learning more broadly to develop more methods that explicitly model the global structure in data.

\section*{Acknowledgements}

The authors thank members of the Harvard DtAK lab for helpful discussions and insights. ASR acknowledges support from the Miami Foundation.  FDV acknowledges support from NSF-CAREER 1750358.

\bibliographystyle{icml2021}
\bibliography{bibliography}

\appendix
\renewcommand{\thesection}{A}
\renewcommand\thefigure{\thesection.\arabic{figure}}
\renewcommand\thetable{\thesection.\arabic{table}}
\setcounter{figure}{0}
\setcounter{table}{0}

\section{Appendix}

\subsection{Training and Architecture Details}\label{sec:training-details}

For Chopsticks, our encoders and decoders used two hidden layers of width 256, and our loss function $\mathcal{L}_x$ was defined as a zero-centered Gaussian negative log likelihood with $\sigma = 0.1$. For Spaceshapes, encoders and decoders used the 7-layer convolutional architecture from \citet{burgess2018understanding}, and our loss function $\mathcal{L}_x$ was Bernoulli negative log likelihood. All models were implemented in Tensorflow; code is available at \url{https://github.com/dtak/hierarchical-disentanglement}.

For both models, the assignment loss $\mathcal{L}_a$ was set to mean-squared error, but only for assignments that were defined: we set undefined assignment components to -1, then let
$\mathcal{L}_a(a,a') = \sum_i\mathbbm{1}[a_i'{\geq}0]  (a_i - a_i')^2$.

All activation functions were set to ReLU ($\max(0,x)$) or Softplus ($\ln(1+e^x)$), e.g. for the initial smooth autoencoder, which was also trained with dimensionality equal to one plus the maximum intrinsic dimensionality of the dataset. We investigate varying this parameter in Fig.~\ref{fig:mimosa-ablations} and find it can be much larger, and perhaps would have produced better results (though nearest neighbor calculation and local SVD computations would have been slower).

All models were trained for 50 epochs with a batch size of 256 on a dataset of size 100,000, split 90\%/10\% into train/test. We used the Adam optimizer with a learning rate starting at 0.001 and decaying by $\frac{1}{10}$ halfway and three-quarters of the way through training.

For COFHAE, we selected softmax temperature $\tau$, the assignment penalty strength $\lambda_1$, and the adversarial penalty strength $\lambda_2$ based on \emph{training set} reconstruction error and MIMOSA assignment accuracy. Splitting off a separate validation set was not necessary, as the most common problem we faced was poor convergence, not overfitting; the adversarial penalty would dominate and prevent the procedure from learning a model that could reconstruct $X$ or $A$. 

Specifically, for each restart, we ran COFHAE with $\tau$ in $\{\frac{1}{2}, \frac{2}{3}, 1\}$, $\lambda_1$ in $\{10,100,1000\}$, and $\lambda_2$ in $\{1,10,100\}$. We then selected the model with the lowest training MSE $\sum_n ||x_n-x_n'||_2^2$,  but restricting ourselves to the 33.3\% of models with the lowest assignment loss $\sum_n \mathcal{L}_a(a_n, a_n')$.

For evaluating $R^4$ and $R^4_c$, we used gradient boosted decision trees, which were faster to train than neural networks.%We also considered using other model-free metrics besides $R^2$ and accuracy, like Spearman rank correlation or normalized mutual information. However, the latter is technically undefined for continuous variables, while the former cannot measure whether categorical variables (with arbitrary order) approximately correspond to continuous variables.

\subsection{Additional Chopsticks Details}\label{sec:more-chopsticks}

In this section, we clarify the generative process behind the different variants of Chopsticks, and discuss alternatives.

Chopsticks can be generated by the following Python code (the exact code we used is slightly different due to the need to save ground-truth factors):

\lstset{language=Python}
\lstset{basicstyle=\footnotesize\ttfamily}
\begin{lstlisting}
def Bern(p):
  return int(np.random.uniform() < p)
    
def Unif(a,b):
  return a + np.random.uniform() * (b-a)

def stick_segment(variant, T):
  slope = Unif(-0.01, 0.01) * np.arange(T)
  inter = Unif( -0.2,  0.2) + np.zeros(T)
  if   variant == 'slope': return slope
  elif variant == 'inter': return inter
  elif variant == 'both': return slope+inter
  elif variant == 'either':
    return slope if Bern(0.5) else inter

def chopsticks(depth, variant, T=64):
  stick = stick_segment(variant, T)
  chop = Bern(1-np.power(2.0,-(depth-1)))
  if chop:
    stick2 = chopsticks(depth-1, variant, T//2)
    stick[T//2:] += stick2
  return stick
\end{lstlisting}

For all variants, at depth $d\geq1$, we sample a linear ``stick'', and then ``chop'' it with probability $1-2^{-(d-1)}$. If we chop the stick, then we recursively generate a new stick of half the length, which we add to the second half of the current stick. We choose chop probabilities in this way so that, on average, we have equal counts of samples at each depth.

\begin{figure}[h]
    \centering

    \subfloat[Chopsticks instances corrupted by Gaussian noise.]{
    \includegraphics[width=0.8\linewidth]{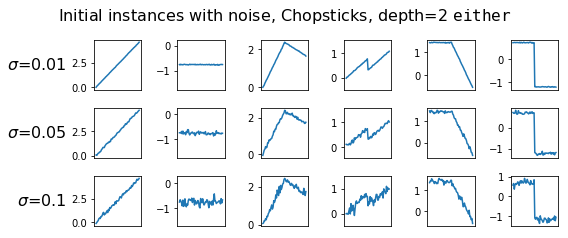}
    \label{fig:mimosa-noise-instances}
    }\\
    \subfloat[Effect of noise on an initial AE representation.]{
    \includegraphics[width=0.75\linewidth]{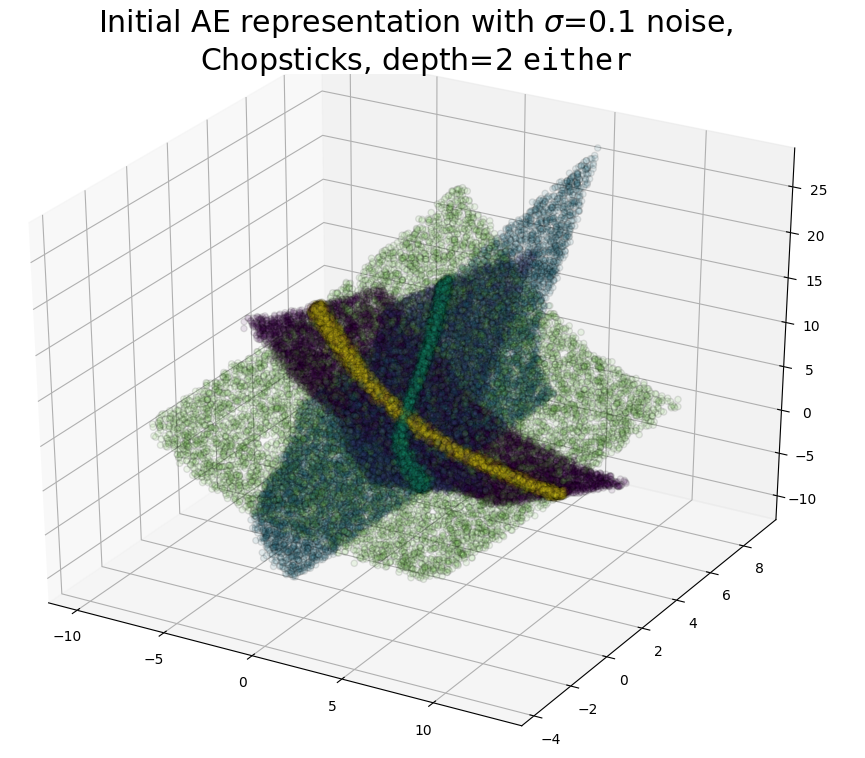}
    \label{fig:mimosa-noise-encoding}
    }\\
    \subfloat[Effect of noise on MIMOSA for two dataset variants.]{
    \includegraphics[width=0.9\linewidth]{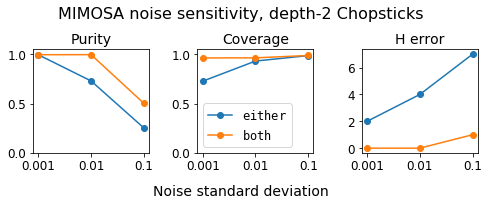}
    \label{fig:mimosa-noise-metrics}
    }
    \caption{Illustration of the sensitivity of MIMOSA to data noise. In preliminary experiments, we find that noise poses the greatest problem for identifying the lowest-dimensional components, e.g. the 1D components in (b) that end up being classified as 2D or 3D. Tuning parameters would help, but we lack labels to cross-validate.}
    \label{fig:mimosa-noise}
\end{figure}

\begin{figure}
    \centering
       \subfloat[Effect of setting a minimum slope/intercept magnitude.]{
    \includegraphics[width=0.49\linewidth]{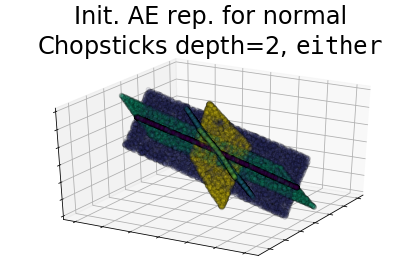}
    \includegraphics[width=0.49\linewidth]{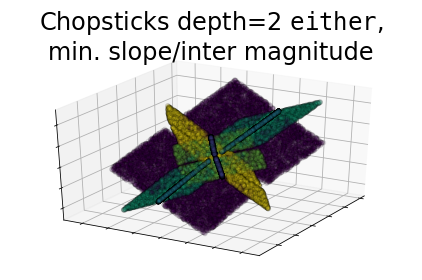}
    \label{fig:chop-gaps}
    } \\
    \subfloat[Effect of overwriting rather than offsetting slope.]{
    \includegraphics[width=0.49\linewidth]{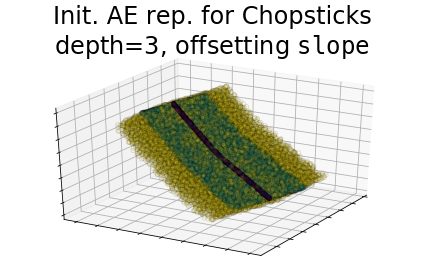}
    \includegraphics[width=0.49\linewidth]{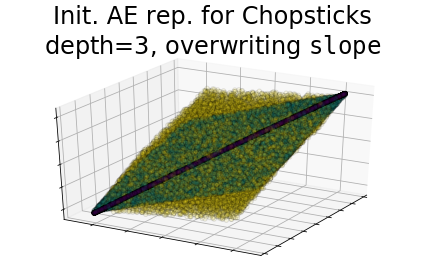}
    \label{fig:chop-sets}
    }
    \caption{Initial AE representations for alternative variants of Chopsticks. Setting a minimum slope/intercept magnitude (a) causes representations to contain gaps.
    Overwriting rather than offsetting slope (b) changes the angle of lower- within higher-dimensional manifolds. Neither change breaks MIMOSA, which can still find the right hierarchy as long as manifolds remain similarly embedded with similar local SVD angles over gaps.}
    \label{fig:chop-vars}
\end{figure}

Although this framework already gives us a wide diversity of datasets, we could consider others. One option is to add noise, which hurts MIMOSA, though it depends on the dataset (Fig.~\ref{fig:mimosa-noise}). Another option is to sample slopes and intercepts non-uniformly or even over non-convex sets; see e.g. Fig.~\ref{fig:chop-gaps}, where we set slope/intercept magnitudes at least a threshold away from 0, which introduces gaps into the initial representation.
In general, MIMOSA continues to return the right hierarchy for all such slope/intercept distributions, though COFHAE disentanglement tends to drop when variables are sampled from Gaussians, likely because of symmetry (with proper rescaling, we can rotate factorized Gaussians in any direction and preserve factorization; the same is not true for uniforms, though \citet{locatello2018challenging} show there must exist analogous, if more complex, transformations).
Yet another possibility is to overwrite the slope and/or intercept when recursing, rather than offsetting them (Fig.~\ref{fig:chop-sets}). Overwriting slope does not affect MIMOSA performance, though it changes the orientation of lower-dimensional within higher-dimensional manifolds, which can affect COFHAE), but overwriting the intercept can break the geometrical nesting of manifolds at large slopes. Although we considered many of these options, we ultimately decided it was pedagogically best for our benchmark to distribute instances over maximally simple (but still arbitrarily deep) underlying manifold structures.

\subsection{Computing $H$-error}\label{sec:h-error-details}

Our $H$-error metric is meant to quantify the ``edit distance'' between two dimension hierarchies $H$ and $\hat{H}$, but in a way that is invariant to merge-up and push-down operations, as well as reorderings of child groups.
To implement it, we first convert $H$ and $\hat{H}$ to a canonical form where each dimension group is labeled by the minimum downstream dimension of its leaves (which equals the dimension of the manifold component at the matching location in the original enclosure hierarchy), which renders us invariant to merge-up and push-down operations. We then reorder children in terms of the (sorted) concatenation of their downstream labels, which renders us invariant to child ordering in most cases.
Finally, we apply the Zhang-Shasha algorithm for tree edit distance between ordered, labeled trees~\citep{zhang1989simple,paassen2015toolbox} to get our final $H$-error.

\subsection{Complexity and Runtimes}\label{sec:runtimes}

\begin{figure}[htb]
    \centering
    \includegraphics[width=\linewidth]{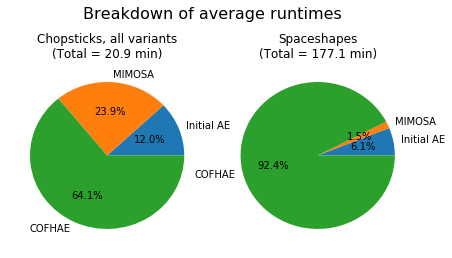}
    \caption{Mean runtimes and percentage breakdowns for COFHAE and MIMOSA on Chopsticks and Spaceshapes, based on Tensorflow implementations running on single GPUs (exact model varies between Tesla K80, Tesla V100, GeForce RTX 2080, etc). Runtimes tend to be dominated by COFHAE, which is similar in complexity to existing adversarial representation learning methods (e.g. FactorVAE).}
    \label{fig:runtimes}
\end{figure}

Per Fig.~\ref{fig:runtimes}, the total runtime of our method is dominated by COFHAE, an adversarial autoencoder method which has the same complexity as FactorVAE~\citep{kim2018disentangling} (linear in dataset size $N$ and number of training epochs, and strongly affected by GPU speed).

MIMOSA could theoretically take more time, however, as the complexity of constructing a ball tree~\citep{omohundro1989five} for nearest neighbor queries is $O(|Z| N \log N)$. As such, initial dimensionality reduction is critical; in our Spaceshapes experiments, $|Z|$ is 7, whereas $|X|$ is 4096.

Other MIMOSA steps can also take time. With a $\mathtt{num\_nearest\_neighbors}$ of $k$, the complexity of running local SVD on every point in the dataset is $O(N(|Z|^2k + |Z|k^2 + k^3))$, providing another reason to reduce initial dimensionality and keep neighborhood size manageable (though ideally $k$ should increase with $|Z|$ to robustly learn local manifold directions). Iterating over the dataset in BuildComponent and computing cosine similarity will also have complexity at least $O(Nkd^3(d+|Z|))$ for components of local dimensionality $d$, and detecting component boundaries can actually have complexity $O(Nke^d)$ (if this is implemented, as in our experiments, by checking if projected points are contained in their neighbors' convex hulls---though we also explored a much cheaper $O(Nk^2d)$ strategy of checking for the presence of neighbors in all principal component directions that worked almost as well).

Although these scaling issues are worth noting, MIMOSA was still relatively fast in our experiments, where runtimes were dominated by neural network training (Fig.~\ref{fig:runtimes}).

%\subsection{Other Metrics}
% FDV: Work this in here 
%\paragraph{Alternative Measures of Correspondence}
%For quantifying correspondence, we considered several alternatives to this symmetrized $R^2$ score, including one based on normalized mutual information ($\frac{I(V_i; Z_j)^2}{H(V_i)H(Z_j)}$) and one based on Spearman rank correlation. However, normalized mutual information is only defined for discrete random variables (which have a well-defined entropy), while Spearman rank correlation is only meaningful when comparing continuous random variables over simply connected domains (where invertibility implies monotonicity). In our experiments, we will be considering mixed discrete and continuous $V$, whose components our method may model with either discrete \emph{or} continuous $Z$.  We observed empirically that our $R^2(V_i \leftrightarrow Z_j)$ score handles all of these cases fairly well.

%TODO

\begin{algorithm*}
\caption{LocalSVD($Z$)}
\label{algo:local-svd}
\begin{algorithmic}[1]
    \STATE Run SVD on $Z$ (a design matrix of dimension \texttt{num\_nearest\_neighbors} by \texttt{initial\_dim})
    
    \IF{$\mathtt{ransac\_frac} < 1$}
    
    \FOR{$\mathrm{each\,dimension}$ $d$ $\mathrm{from}$ $1$ $\mathrm{to}$ $\mathtt{initial\_dim}-1$}
    
    \FOR{$\mathrm{each\,point}\, z_n$}
            
    \STATE Compute the reconstruction error for $z_n$ using the only first $d$ SVD dimensions
    
    \ENDFOR
    
    \ENDFOR
    
    \STATE Take the norm of reconstruction errors across dimensions, giving a vector of length $\mathtt{num\_nearest\_neighbors}$
        
    \STATE Re-fit SVD on points whose error-norms are less than the $100 \times \mathtt{ransac\_frac}$ percentile value.
    
    \ENDIF
    
    \FOR{$\mathrm{each\,dimension}$ $d$ $\mathrm{from}$ $1$ $\mathrm{to}$ $\mathtt{initial\_dim}-1$}
    
    \STATE Check if the cumulative sum of the first $d$ eigenvalues is at least \texttt{eig\_cumsum\_thresh}
        
    \STATE Check if the ratio of the $d$th to the $d+1$st eigenvalue is at least \texttt{eig\_decay\_thresh}
        
    \IF{$\mathrm{both\,of\,these\,conditions\,are\,true}$}
        \STATE \textbf{return} only the first $d$ SVD components
    \ENDIF    
    
    \ENDFOR

    \STATE \textbf{return} the full set of SVD components otherwise

\end{algorithmic}
\end{algorithm*}

\begin{algorithm*}
\caption{TangentPlaneCos($U,V$)}
\label{algo:cos}
\begin{algorithmic}[1]
    \IF{$U$ and $V$ are equal-dimensional}
   
    \STATE  \textbf{return} $|\det(U \cdot V^T)|$
    
    \ELSE
    
    \STATE \textbf{return} 0
    
    \ENDIF
\end{algorithmic}
\end{algorithm*}

\begin{algorithm*}
\caption{BuildComponent($z_i$, neighbors, svds)}
\label{algo:build-component}
\begin{algorithmic}[1]
    
    \STATE Initialize component to $z_i$ and neighbors $z_j$ not already in other components where $\mathrm{TangentPlaneCos}(\mathrm{svds}_i, \mathrm{svds}_j) \geq \mathtt{cos\_simil\_thresh}$ (Algorithm \ref{algo:cos}).
    
    \WHILE{the component is still growing}
        \STATE Add all points $z_k$ for which at least \texttt{contagion\_num} of their neighbors $z_{\ell}$ are already in the component with $\mathrm{TangentPlaneCos}(\mathrm{svds}_k, \mathrm{svds}_\ell) \geq \mathtt{cos\_simil\_thresh}$.
        
        \STATE Skip adding any $z_k$ already in another component.
    \ENDWHILE
    
    \STATE \textbf{return} the set of points in the component
    
    %\STATE \textcolor{gray}{\# Note that this is run repeatedly over the full dataset, starting at $z_1$, and keeping track of the set of points already added so they can be skipped.}
\end{algorithmic}
\end{algorithm*}

\begin{algorithm*}
\caption{MergeComponents(components, svds)}
\label{algo:merge-components}
\begin{algorithmic}[1]
    
    \STATE Discard components smaller than $\mathtt{min\_size\_init}$.
    
    \FOR{each component $c_i$}
       
    \STATE Construct a local ball tree for the points in $c_i$.
        
    \STATE Set $c_i$.edges to points not contained in the convex hull of their neighbors in local SVD space.
    
    \ENDFOR
    
    \STATE Initialize edge overlap matrix $M$ of size $|$components$|$ by $|$components$|$ to 0.
    
    \FOR{each ordered pair of equal-dimensional components $(c_i, c_j)$}
    
    \STATE  Set $M_{ij}$ to the fraction of points in $c_i$.edges for which the closest point in $c_j$.edges has local SVD tangent plane similarity above \texttt{cos\_simil\_thresh}.
    
    \ENDFOR
    
    \STATE Average $M$ with its transpose to symmetrize.
    
    \STATE Merge all components $c_i \neq c_j$ of equal dimensionality $d$ where $M_{ij} \geq \mathtt{min\_common\_edge\_frac}(d)$.
    
    \STATE Discard components smaller than $\mathtt{min\_size\_merged}$.
    
    \STATE \textbf{return} the merged set of components
\end{algorithmic}
\end{algorithm*}

\begin{algorithm*}
\caption{ConstructHierarchy(components)}
\label{algo:construct-hier}
\begin{algorithmic}[1]
    \FOR{each component $c_i$}
       \STATE Set $c_i$.neighbor\_lengthscale to the average distance of points to their nearest neighbors inside the component (computed using the local ball tree from Algorithm~\ref{algo:merge-components})
    \ENDFOR

    \FOR{each pair of different-dimensional components $(c_i, c_j)$, $c_i$ higher-dimensional}
        \STATE Compute the average distance from points in $c_i$ to their nearest neighbors in $c_j$ (via ball tree).
        
        \STATE Divide this average distance by $c_i.\mathrm{neighbor\_lengthscale}$.
        
        \IF{$\mathrm{the\,resulting\,ratio} \leq \mathtt{neighbor\_lengthscale\_mult}$}
        
            \STATE Set $c_j \in c_i$ ($c_j$ is enclosed by $c_i$)
        \ENDIF
    \ENDFOR
    
    \STATE Create a root node with edges to all components which do not enclose others.

    \STATE Transform the component enclosure DAG into a tree (where enclosing components are children of enclosed components) by deleting edges which: \begin{enumerate}
      \itemsep0em 
        \item are redundant because an intermediate edge exists, e.g. if $c_1 \in c_2 \in c_3$, we delete the edge between $c_1$ and $c_3$.
        \item are ambiguous because a higher-dimensional component encloses multiple lower-dimensional components (i.e. has multiple parents). In that case, preserve only the edge with the lowest distance ratio.% (variant 1, our default), or move the edge further up in the hierarchy to the earliest common ancestor (variant 2, another option we considered).
    \end{enumerate}
    
    \STATE Convert the resulting component enclosure tree into a dimension hierarchy:
    \begin{enumerate}
    \itemsep0em 
        \item If the root node has only one child, set it to be the root. Otherwise, begin with a dimension group with a single categorical dimension whose options point to groups containing each child.
        \item For the rest of the component tree, add continuous dimensions until the total number of continuous dimensions up to the root equals the component's dimensionality.
        \item If a component has children, add a categorical dimension that includes those child groups as options (recursing down the tree), along with an empty group (\fbox{$\emptyset$}) option. 
    \end{enumerate}
    
    \STATE \textbf{return} the dimension hierarchy
\end{algorithmic}
\end{algorithm*}

\begin{algorithm*}
\caption{$\mathrm{HAE}_\theta$.encode($x; \tau$)}
\label{algo:encode}
\begin{algorithmic}[1]
    \STATE Encode $x$ using any neural network architecture as a flat vector $z_{pre}$, with size equal to the number of continuous variables plus the number of categorical options in $\mathrm{HAE}_{\theta}.\mathrm{hierarchy}$.
    
    \STATE Associate each group of dimensions in the flat vector with variables in the hierarchy.
    
    \STATE For all of the categorical variables, pass their options through a softmax with temperature $\tau$.
    
    \STATE Use the softmax outputs to recursively mask all components of $z_{pre}$ corresponding to variables \emph{below} each option in $\mathrm{HAE}_{\theta}.\mathrm{hierarchy}$.

    \STATE \textbf{return} the masked representation, separated into discrete $a'$, continuous $z$, as well as the mask $m$ (for determining active dimensions later).
\end{algorithmic}
\end{algorithm*}

\begin{figure*}
    \centering
    \includegraphics[width=0.45\linewidth]{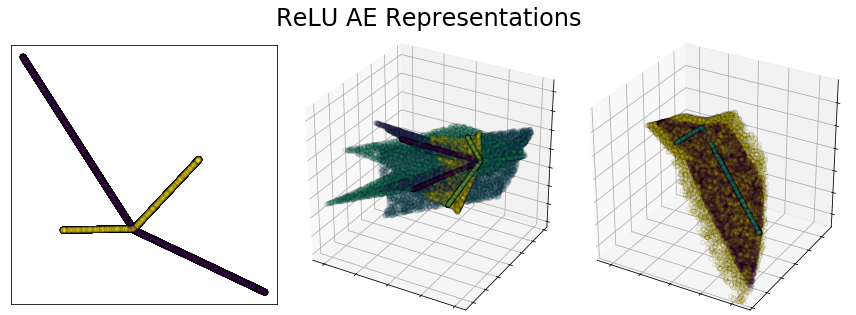}
    \hspace{1cm}
    \includegraphics[width=0.45\linewidth]{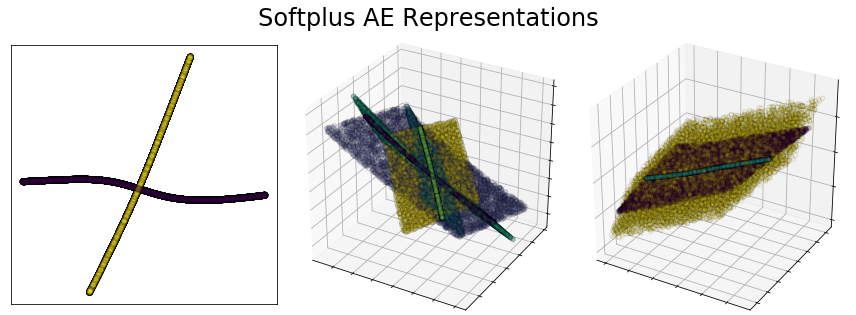}
    
    \caption{Comparison of the latent spaces learned by MIMOSA initial autoencoders with ReLU (left) vs. Softplus (right) activations on three versions of Chopsticks (depth=1 $\mathtt{either}$, depth=2 $\mathtt{either}$, and depth=3 $\mathtt{slope}$). Each plot shows encoded data samples colored by their ground-truth location in the dimension hierarchy. Because ReLU activations are non-differentiable at $0$, the resulting latent manifolds contain sharp corners where local SVD directions change discontinuously, causing issues for BuildComponent and MergeComponents within MIMOSA (Algorithms~\ref{algo:build-component} and \ref{algo:merge-components}). Representations learned by autoencoders with smooth activation functions work much better.}
    \label{fig:relu-vs-softplus}
\end{figure*}

\subsection{MIMOSA Hyperparameters}\label{sec:mimosa-hypers}

\begin{figure*}
    \centering
    \includegraphics[width=\textwidth]{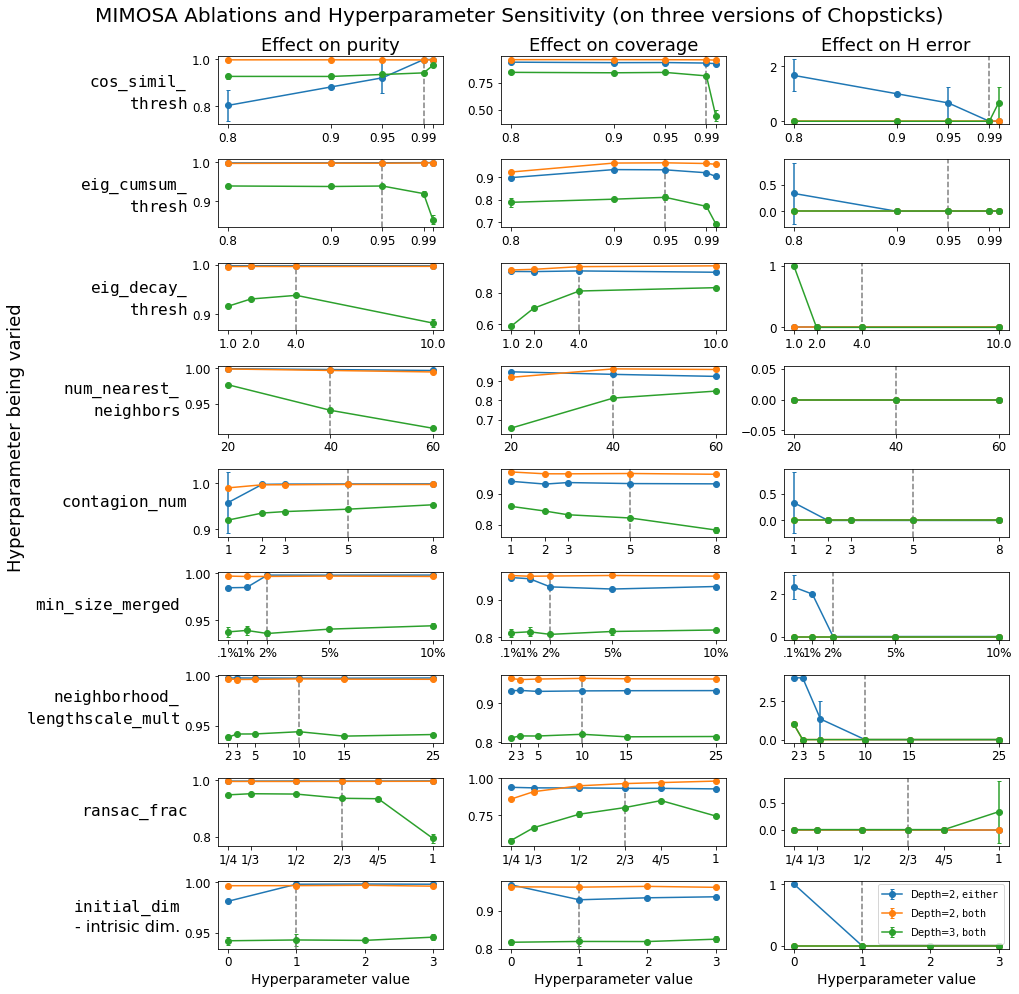}
    \caption{Effect of varying different hyperparameters (and ablating different robustness techniques) on MIMOSA. Default values are shown with vertical gray dotted lines, and for each hyperparameter (top to bottom), average coverage (left), purity (middle), and $H$ error (right) when deviating from defaults are shown for three versions of the Chopsticks dataset. Results suggest both a degree of robustness to changes (degradations tend not to be severe for small changes), but also the usefulness of various components; for example, results markedly improve on some datasets with $\mathtt{contagion\_num}{>}1$ and $\mathtt{ransac\_frac}{<}1$ (implying contagion dynamics and RANSAC both help). Many parameters exhibit tradeoffs between component purity and dataset coverage.}
    \label{fig:mimosa-ablations}
\end{figure*}

In this section, we list and describe all hyperparameters for MIMOSA, along with values that we used for our main results. We also present sensitivity analyses in Fig.~\ref{fig:mimosa-ablations}.

\textbf{MIMOSA initial autoencoder (Algorithm~\ref{algo:mimosa}, line 1)} 
        \begin{itemize}[topsep=0pt]
        \itemsep0em 
            \item \texttt{initial\_dim} - the dimensionality of the initial smooth autoencoder. As the sensitivity analysis in Fig.~\ref{fig:mimosa-ablations} shows, this does not need to be as low as the intrinsic dimensionality of the data, which MIMOSA will estimate, and ideally should be a little larger. We defaulted to using the maximum intrinsic dimensionality plus 1; in a real-world context where this information is not available, it can be estimated by starting at $\mathtt{initial\_dim} = |X|$ and reducing until initial autoencoder reconstruction error starts increasing.
            \item Training and architectural details appropriate for the data modality (e.g. convolutional layers for images). See \S\ref{sec:training-details} for our choices.
        \end{itemize}
\textbf{LocalSVD (Algorithm~\ref{algo:local-svd})} 
       \begin{itemize}[topsep=0pt]
       \itemsep0em 
          \item \texttt{num\_nearest\_neighbors} - neighborhood size for LocalSVD and traversal; we used $40$. Must exceed $\texttt{initial\_dim}$; could replace with a search radius.
          \item \texttt{ransac\_frac} - the fraction of neighbors to refit SVD. We used $2/3$. Note that we do not run traditional, multi-step RANSAC~\citep{fischler1981random}, but a more efficient two-step approximation, where we define the loss term based an aggregation of reconstruction errors across dimensions. Another (less efficient but potentially more robust) option would be to iteratively re-fit SVD using the points with lowest reconstruction error at each dimension, and check if the resulting eigenvalues meet our cutoff criteria.
          \item \texttt{eig\_cumsum\_thresh} - the minimum fraction of variance SVD dimensions must explain to determine local dimensionality. We used $0.95$. For noisy or sparse data, it might be useful to reduce this parameter.
          \item \texttt{eig\_decay\_thresh} - the minimum multiplicative factor by which SVD eigenvalues must decay to determine local dimensionality. We used $4$. It might also be useful to reduce this parameter for sparse data.
        \end{itemize}
Note that our LocalSVD algorithm can be seen as a faster version of Multiscale SVD~\citep{little2009estimation}, which is used in an analogous way by \citet{mahapatra2017s}, but would require repeatedly computing singular value decompositions over different search radii for each point.
 
\textbf{BuildComponent (Algorithm~\ref{algo:build-component})} 
        \begin{itemize}[topsep=0pt]
       \itemsep0em 
            \item \texttt{cos\_simil\_thresh} - neighbors' local SVDs must be this similar to add to the component. This corresponds to the $\epsilon$ parameter from \citet{mahapatra2017s}. We used $0.99$ for Chopsticks and $0.95$ for Spaceshapes; in general, we feel this is one of the most important parameters to tune, and should generally be reduced in the presence of noise or data scarcity.
            \item \texttt{contagion\_num} - only add similar points to a manifold component when a threshold fraction of their neighbors have already been added. This is useful for robustness, and corresponds to the $T$ parameter from \citet{mahler2020contagion} (but expressed as a number rather than a fraction). We used $5$ for Chopsticks and $3$ for Spaceshapes. Values above ~$20\%$ of $\mathtt{num\_nearest\_neighbors}$ will likely produce poor results, and we found the greatest increases in robustness just going from 1 (or no contagion dynamics) to 2.
        \end{itemize}
  
\textbf{MergeComponents (Algorithm~\ref{algo:merge-components})}
    \begin{itemize}[topsep=0pt]
       \itemsep0em 
        \item \texttt{min\_size\_init} - discard initial components smaller than this, which helps speed up the algorithm (by reducing the number of pairwise comparisons) and avoid merges through single-point components. We used $0.02\%$ of the dataset size, or 20 points. 
        \item \texttt{min\_size\_merged} - discard merged components smaller than this, which helps exclude spurious interstitial points at boundaries where low-dimensional components intersect. We used $2\%$ of the dataset size, or 2000 points.
        \item \texttt{min\_common\_edge\_frac(d)} - the minimum fraction of edges that two manifold components must share in common to merge, as a function of dimensionality $d$. We used $2^{-d-1} + 2^{-d-2}$; this is based on the idea that two neighboring (possibly distorted) hypercubes of dimension $d$ should match on one of their sides; since they have $2^d$ sides, the fraction of matching edge points would be $2^{-d}$. However, for robustness (as not all manifold segments will be hypercubes, and even then some edge points may not match), we average that fraction with the smaller fraction that would need for a $d+1$ dimensional hypercube, or $2^{-d-1}$, for our resulting $2^{-d-1} + 2^{-d-2}$. In general, we found that this choice was not critical in the noiseless data case, as matches were common for separated components with the same true assignments and rare for others, but it did help in cases with many intersecting components.% and may be more important with noise.
    \end{itemize}

\textbf{ConstructHierarchy (Algorithm~\ref{algo:construct-hier})}
\begin{itemize}[topsep=0pt]
       \itemsep0em 
    \item \texttt{neighbor\_lengthscale\_mult} - the threshold for deciding whether a higher-dimensional component ``encloses'' a lower-dimensional component, expressed as a ratio of (1) the average distance from lower-dimensional component points to their nearest neighbors in the higher-dimensional component (inter-component distance), to (2) the average distance of points in the higher-dimensional component to their nearest neighbors in that same component (intra-component distance). We used $10$, which we found was robust for our benchmarks, though it may need to be increased if ground-truth components are higher-dimensional than those in our benchmarks.
\end{itemize}

\begin{figure*}
    \centering

    \includegraphics[width=0.9\textwidth]{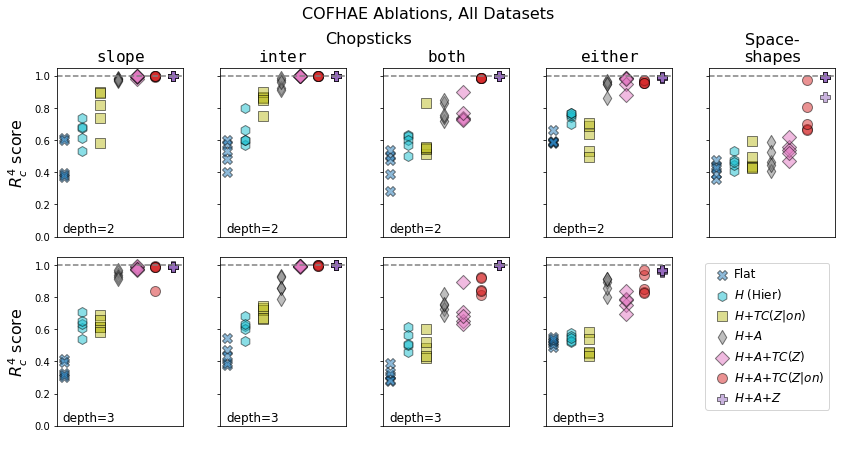}
    \caption{A fuller version of main paper Fig.~\ref{fig:cofhae-ablations} showing COFHAE ablations on all datasets. Hierarchical disentanglement tends to be low for flat AEs (Flat), better with ground-truth hierarchy $H$ (Hier $H$), and even better after adding supervision for ground-truth assignments $A$ ($H{+}A$). Adding a FactorVAE-style marginal TC penalty ($H{+}A{+}TC(Z)$) sometimes helps disentanglement, but making that TC penalty conditional ($H{+}A{+}TC(Z|on)$, i.e. COFHAE) tends to help more, bringing it close to the near-optimal disentanglement of a hierarchical model whose latent representation is fully supervised ($H{+}A{+}Z$).
    Partial exceptions include the hardest three datasets (Spaceshapes and depth-3 compound Chopsticks), where disentanglement is not consistently near 1; this may be due to non-identifiability or adversarial optimization difficulties.
    }
    \label{fig:ablations-all}
\end{figure*}

\begin{figure*}[h]
    \centering
    \includegraphics[width=\linewidth]{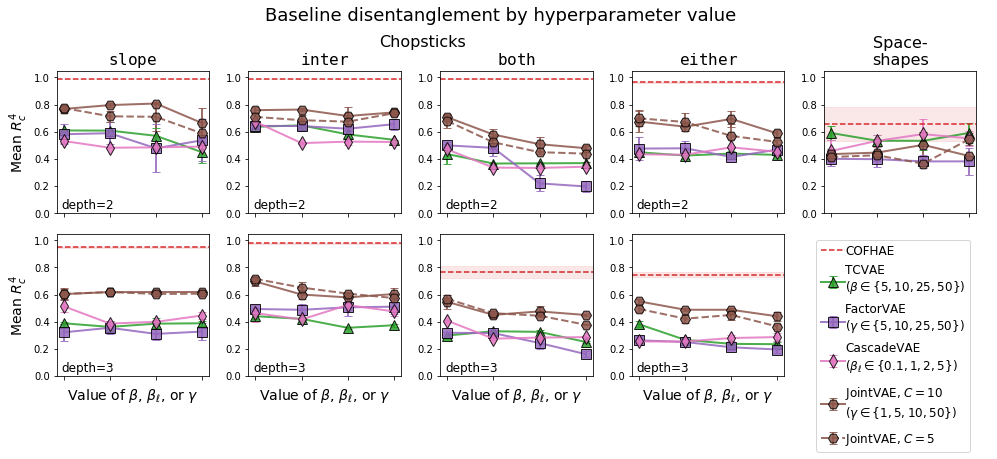}
    \caption{Varying disentanglement penalty hyperparameters for baseline algorithms (TCVAE, FactorVAE, CascadeVAE, and JointVAE). Markers indicate mean $R^4_c$ over 5 trials, with standard deviation errorbars. In contrast to COFHAE (mean performance in red, with standard deviation in pink), no setting produces near-optimal disentanglement, and disentanglement often \emph{decreases} with increasing disentanglement penalty strength.}
    \label{fig:baseline-hypers}
\end{figure*}

\begin{figure*}
    \centering
    
    \subfloat[AE pairwise histograms and $R^4$/$R^4_c$ scores]{
        \includegraphics[width=0.49\textwidth]{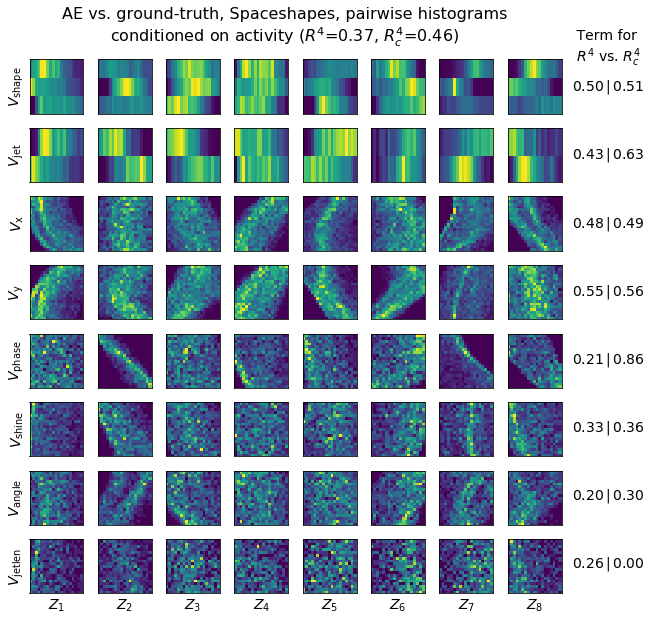}
        \label{fig:ae-spaceshapes-qual}
    }
    \subfloat[TCVAE pairwise histograms and $R^4$/$R^4_c$ scores]{
    \includegraphics[width=0.49\textwidth]{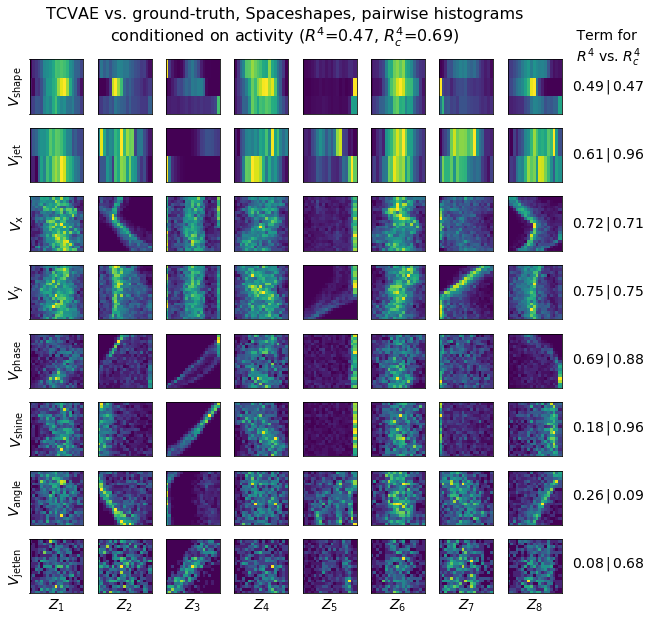}
    \label{fig:tcvae-spaceshapes-qual}
    }
    
    \subfloat[COFHAE pairwise histograms and $R^4$/$R^4_c$ scores]{
    \includegraphics[width=\textwidth]{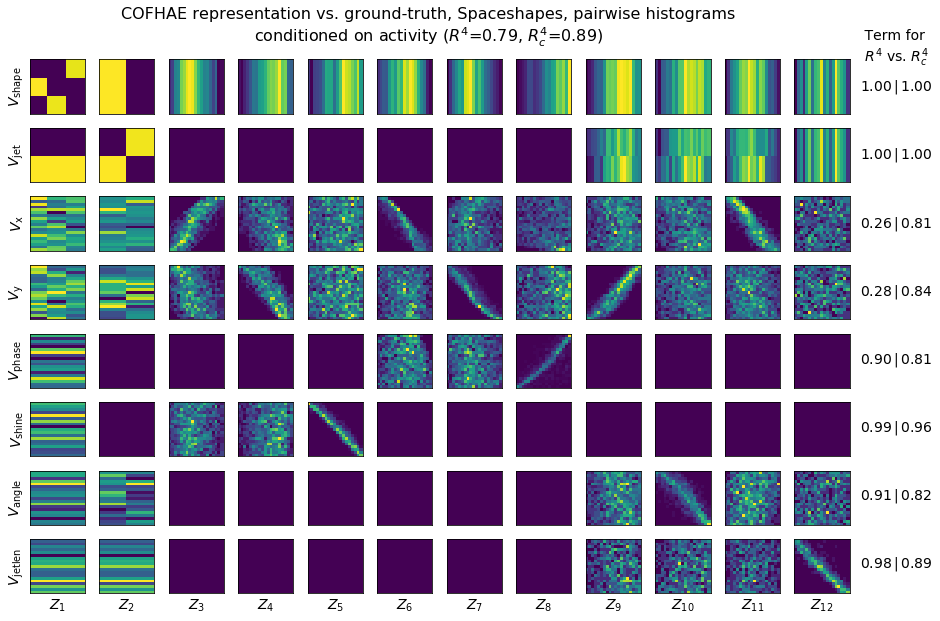}
    \label{fig:cofhae-spaceshapes-qual}
    }
    \caption{Pairwise histograms of ground-truth vs. learned variables for a flat autoencoder (top left), $\beta$-TCVAE (top right), and the best-performing run of COFHAE (bottom) on Spaceshapes. Histograms are conditioned on both variables being active, and dimension-wise components of the $R^4_c$ score are shown on the right. $\beta$-TCVAE does a markedly better job disentangling certain components than the flat autoencoder, but in this case, COFHAE is able to fully disentangle the ground-truth by modeling the discrete hierarchical structure. See Fig.~\ref{fig:hier-latent-traversal} for a hierarchical latent traversal, or \url{https://hreps.s3.amazonaws.com/viz/index.html?dataset=spaceshapes&model=cofhae} for an interactive visualization.}
    \label{fig:cofhae-spaceshapes-qualitative}
\end{figure*}

\begin{figure*}
\centering
\includegraphics[width=\textwidth]{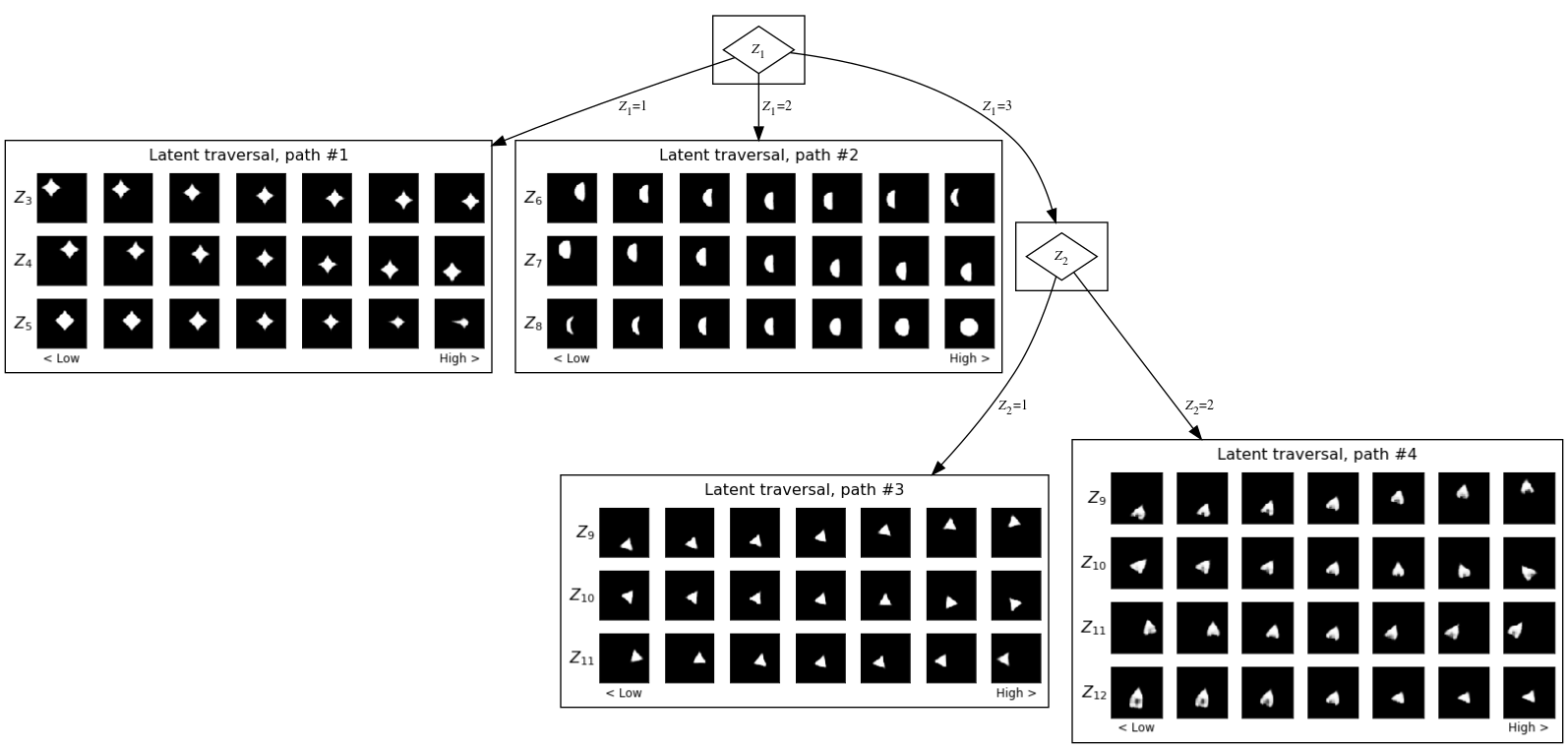}

\caption{Hierarchical latent traversal plot for the Spaceshapes COFHAE model shown in Fig.~\ref{fig:cofhae-spaceshapes-qual}. Individual latent traversals show the effects of linearly sweeping each \emph{active} dimension from its 1st to 99th percentile value (center column shows the same input with intermediate values for all active dimensions). Consistent with Fig.~\ref{fig:cofhae-spaceshapes-qual}, the model is not perfectly disentangled, though primary correspondences are clear: star $\mathtt{shine}$ is modeled by $Z_5$, moon $\mathtt{phase}$ is modeled by $Z_8$, ship $\mathtt{angle}$ is modeled by $Z_{10}$, ship $\mathtt{jetlen}$ is modeled by $Z_{12}$, and $(\mathtt{x},\mathtt{y})$ are modeled by $(Z_{3},Z_{4})$, $(Z_{6},Z_{7})$, and $(Z_{11},Z_{9})$ respectively for each shape. See also an \href{https://hreps.s3.amazonaws.com/viz/index.html?dataset=spaceshapes&model=cofhae}{interactive visualization}.}
\label{fig:hier-latent-traversal}
\end{figure*}

\begin{figure*}
\fbox{\includegraphics[width=0.255\textwidth]{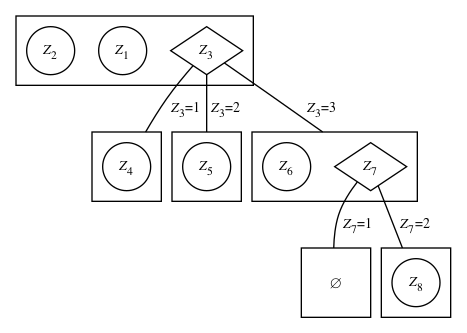}}
\fbox{\includegraphics[width=0.255\textwidth]{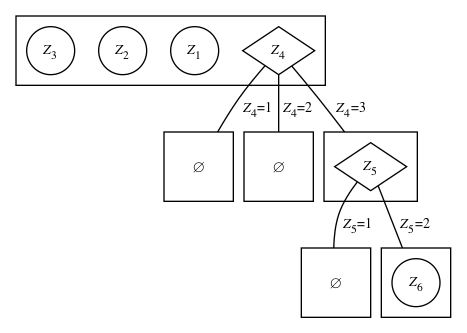}}
\fbox{\includegraphics[width=0.439\textwidth]{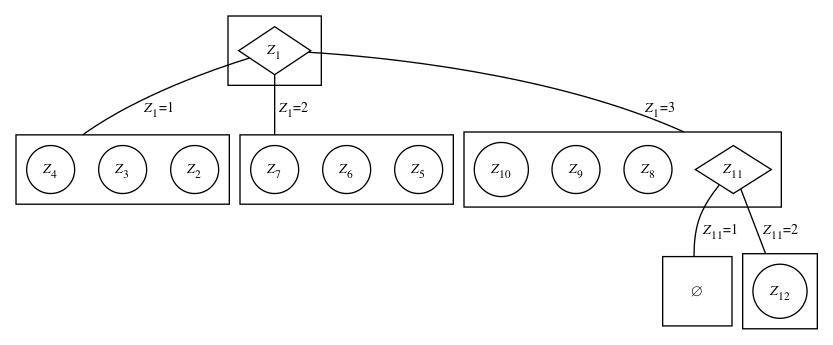}}
\caption{Three different potential hierarchies for Spaceshapes which all have the same structure of variable groups and dimensionalities, but with different distributions of continuous variables across groups.
The ambiguity in this case is that the continuous variable that modifies each shape ($\mathtt{phase}, \mathtt{shine}, \mathtt{angle}$) could either be a child of the corresponding shape category, or be ``merged up'' and combined into a single top-level continuous variable that controls the shape in different ways based on the category.  Alternatively, the location variables $\mathtt{x}$ and $\mathtt{y}$ could instead be ``pushed down'' from the top level and duplicated across each shape category.  In each of these cases, the learned representation still arguably disentangles the ground-truth factors---in the sense that for any fixed categorical assignment, there is still 1:1 correspondence between all learned and ground-truth continuous factors. 
We deliberately design our $R^4_c$ and $H$-error metrics in \S\ref{sec:metrics} to be invariant to these transformations, leaving this specific disambiguation to future work.}
\label{fig:spaceshapes-alts}
\end{figure*}

\begin{figure*}
\centering

\resizebox{\textwidth}{!}{%
\includegraphics[height=3cm]{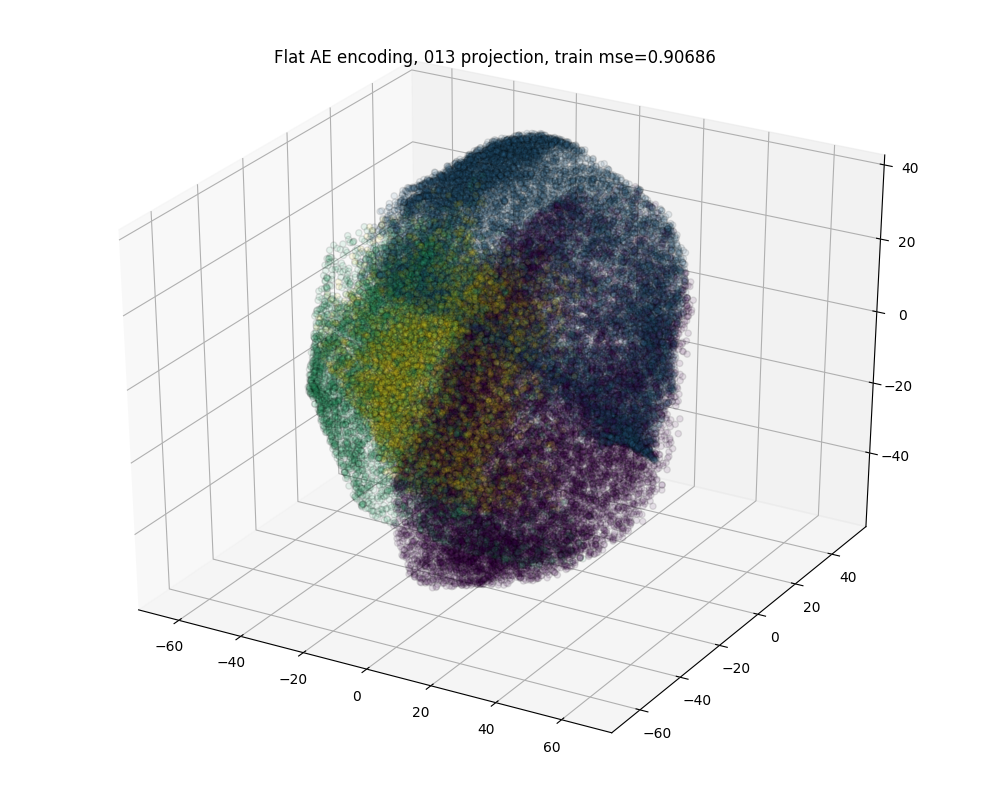}
\includegraphics[height=3cm]{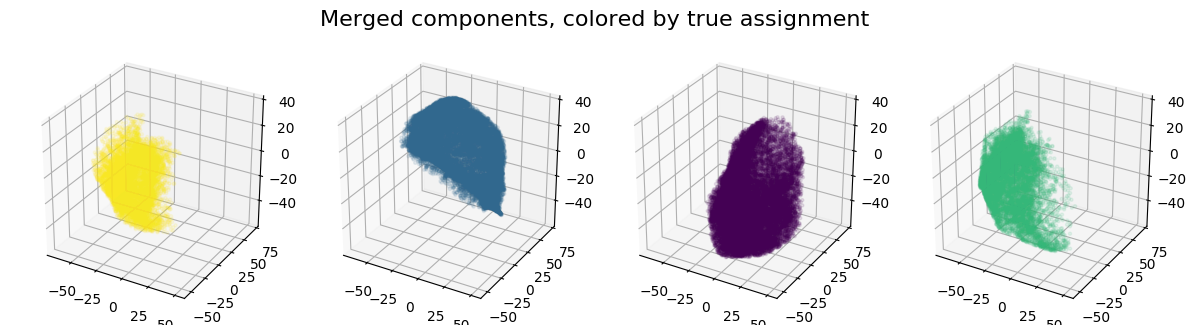}
\includegraphics[height=3cm]{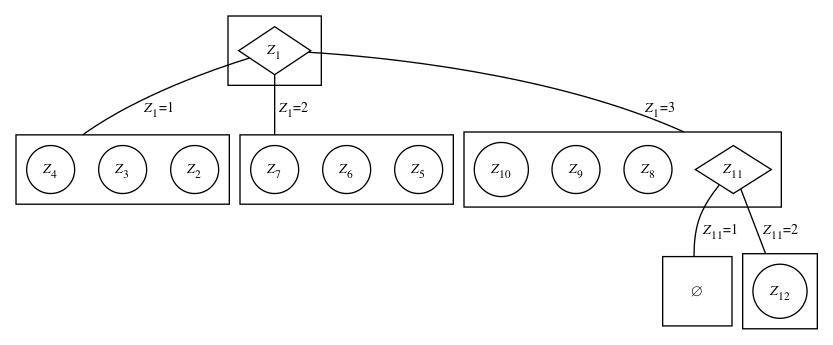}
}
\caption{MIMOSA-learned initial encoding (left), components (middle), and hierarchy (right) for Spaceshapes. Initial points are in 7 dimensions and projected to 3D for plotting. Three identified components are 3D and one is 4D. Analogue of Fig.~\ref{fig:chopsticks-manifolds} in the main text.}
\label{fig:spaceshapes-manifolds}
\end{figure*}

\begin{figure*}
\centering
\resizebox{\textwidth}{!}{%
\includegraphics[height=3cm]{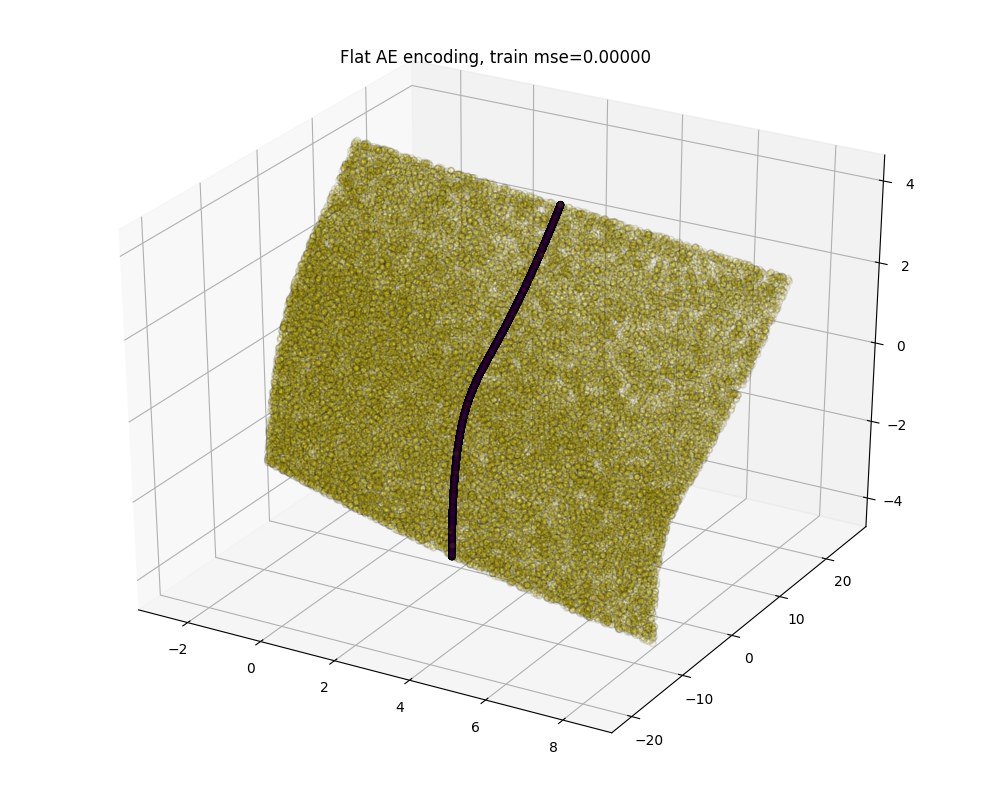}
\includegraphics[height=3cm]{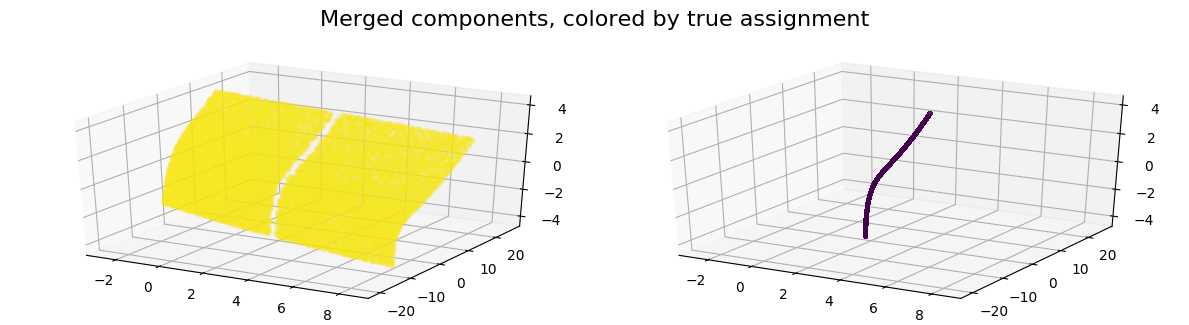}
\includegraphics[height=3cm]{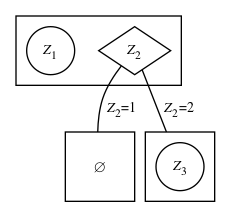}
}
\caption{MIMOSA-learned initial encoding (left), 2D and 1D components (middle), and hierarchy (right) for depth-2 Chopsticks varying the slope. Analogue of Fig.~\ref{fig:chopsticks-manifolds} in the main text.}
\label{fig:chopsticks201-manifolds}
\end{figure*}

\begin{figure*}
\centering
\resizebox{\textwidth}{!}{%
\includegraphics[height=3cm]{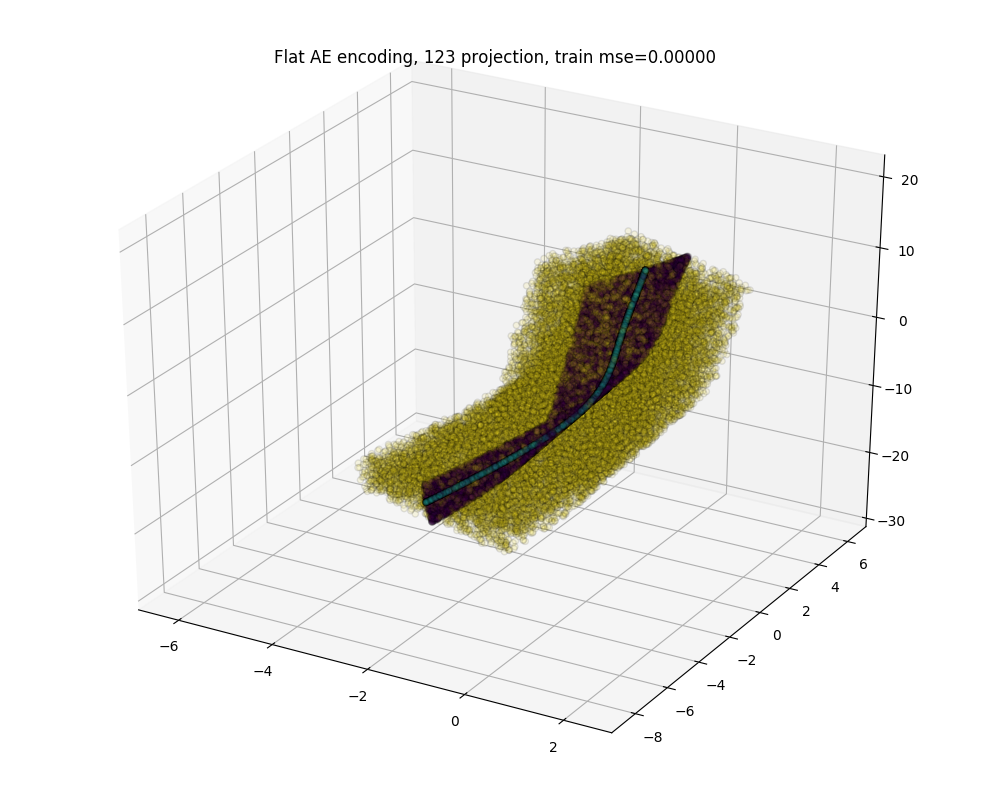}
\includegraphics[height=3cm]{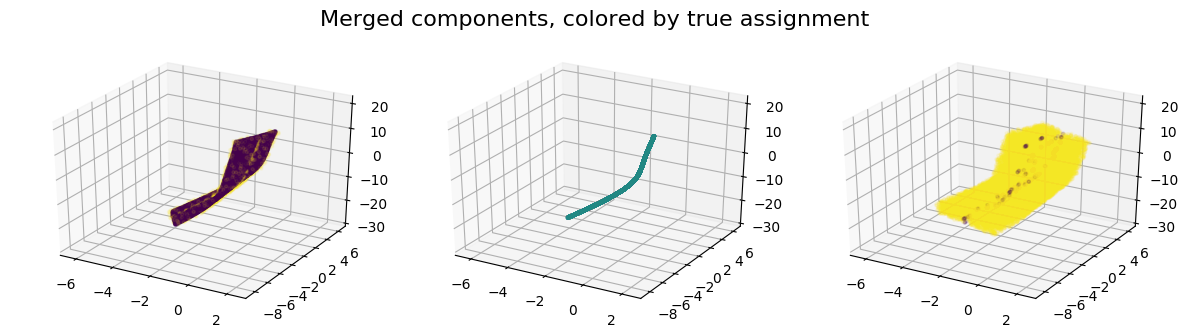}
\includegraphics[height=3cm]{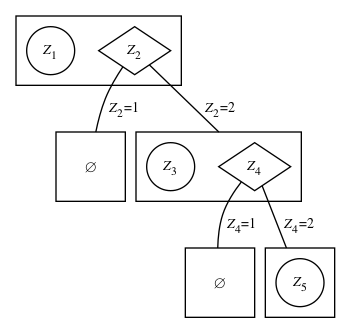}
}
\caption{MIMOSA-learned initial encoding (left), 2D, 1D, and 3D components (middle), and hierarchy (right) for depth-3 Chopsticks varying the slope. Initial points are in 4 dimensions and projected to 3D for plotting. Analogue of Fig.~\ref{fig:chopsticks-manifolds} in the main text.}
\label{fig:chopsticks301-manifolds}
\end{figure*}

\begin{figure*}
\centering
\resizebox{\textwidth}{!}{%
\includegraphics[height=3cm]{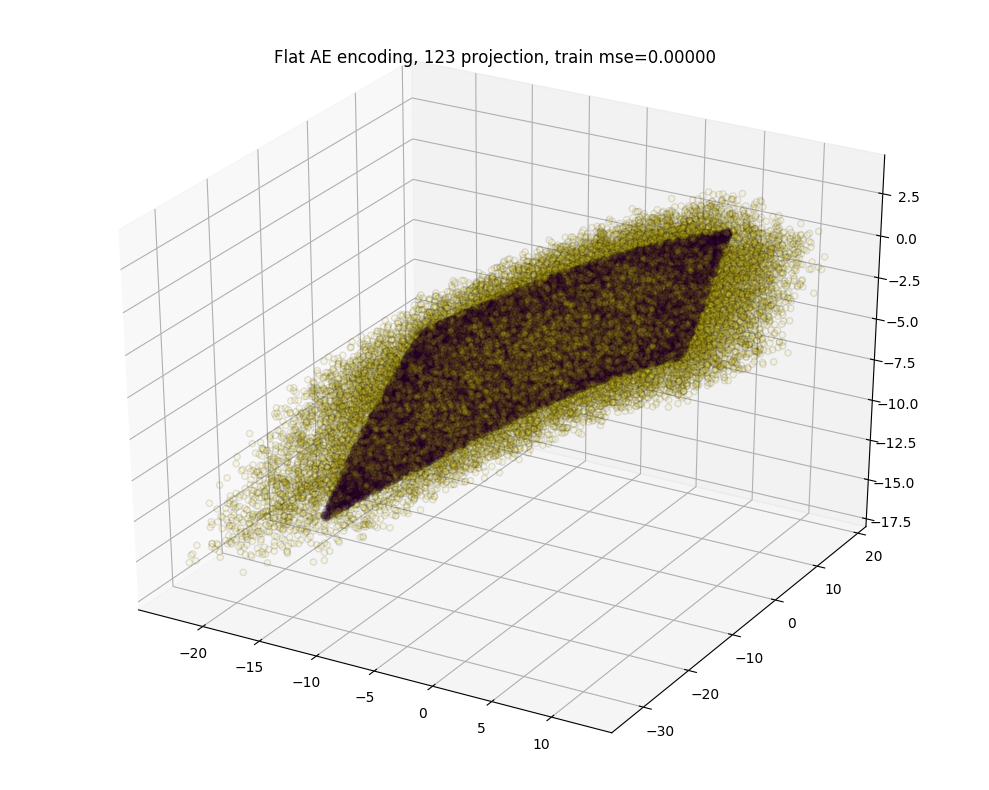}
\includegraphics[height=3cm]{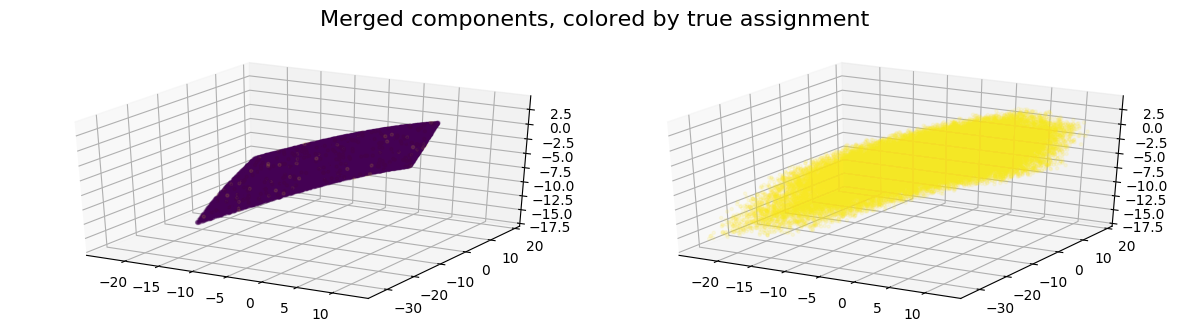}
\includegraphics[height=3cm]{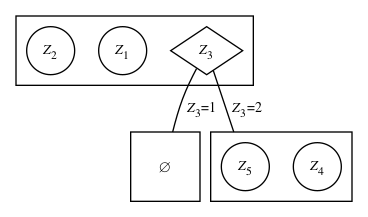}
}
\caption{MIMOSA-learned initial encoding (left), 2D and 4D components (middle), and hierarchy (right) for depth-2 Chopsticks varying $\mathtt{both}$ slope and intercept. Initial points are in 5 dimensions and projected to 3D for plotting. Analogue of Fig.~\ref{fig:chopsticks-manifolds} in the main text.}
\label{fig:chopsticks211-manifolds}
\end{figure*}

\begin{figure*}
\centering
\resizebox{\textwidth}{!}{%
\includegraphics[height=3cm]{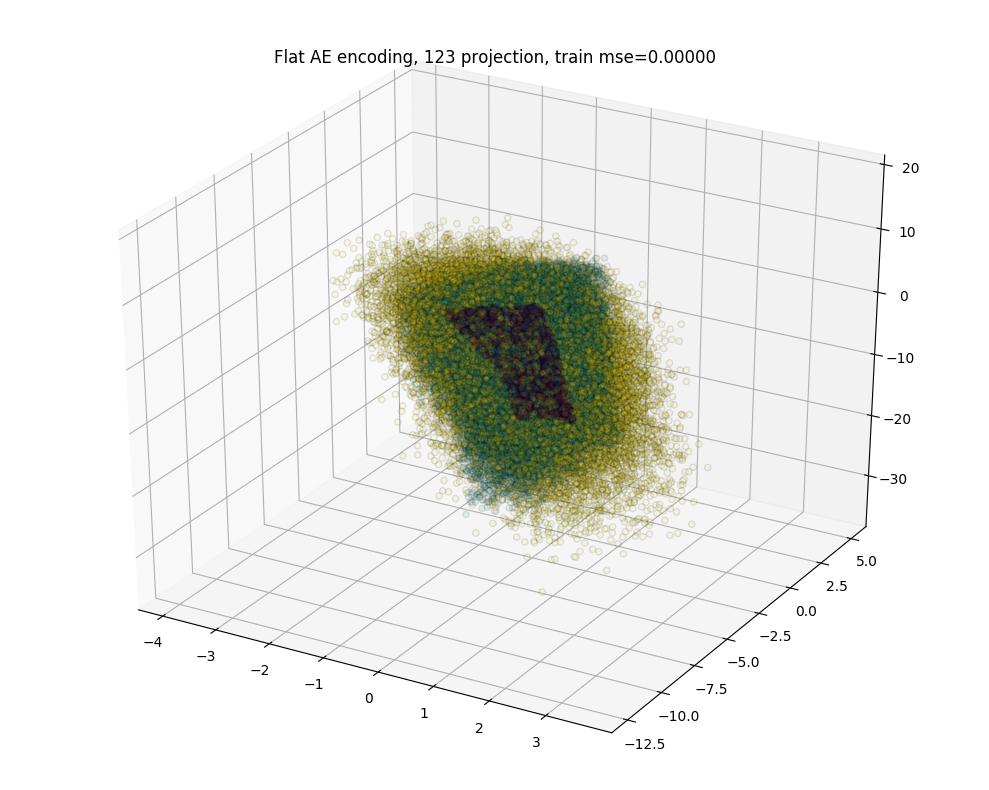}
\includegraphics[height=3cm]{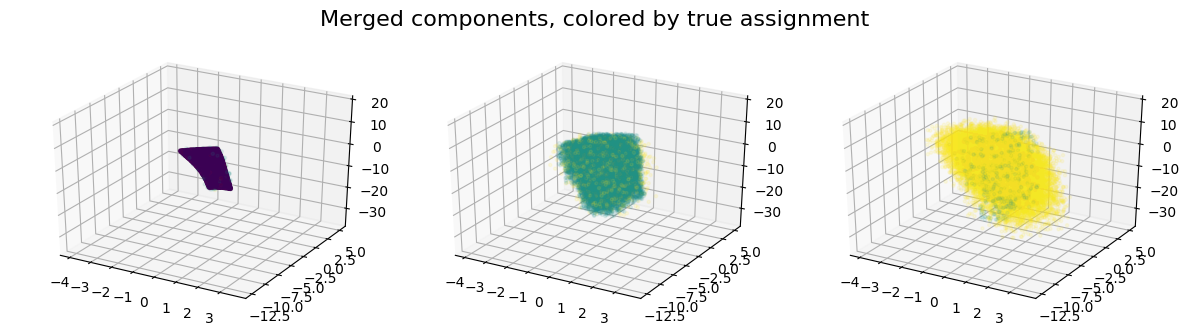}
\includegraphics[height=3cm]{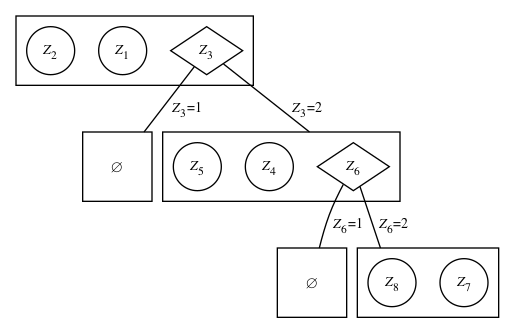}
}
\caption{MIMOSA-learned initial encoding (left), 2D, 4D, and 6D  components (middle), and hierarchy (right) for depth-2 Chopsticks varying $\mathtt{both}$ slope and intercept. Initial points are in 7 dimensions and projected to 3D for plotting. Analogue of Fig.~\ref{fig:chopsticks-manifolds} in the main text.}
\label{fig:chopsticks311-manifolds}
\end{figure*}

\begin{figure*}
\centering
\resizebox{\textwidth}{!}{%
\includegraphics[height=3cm]{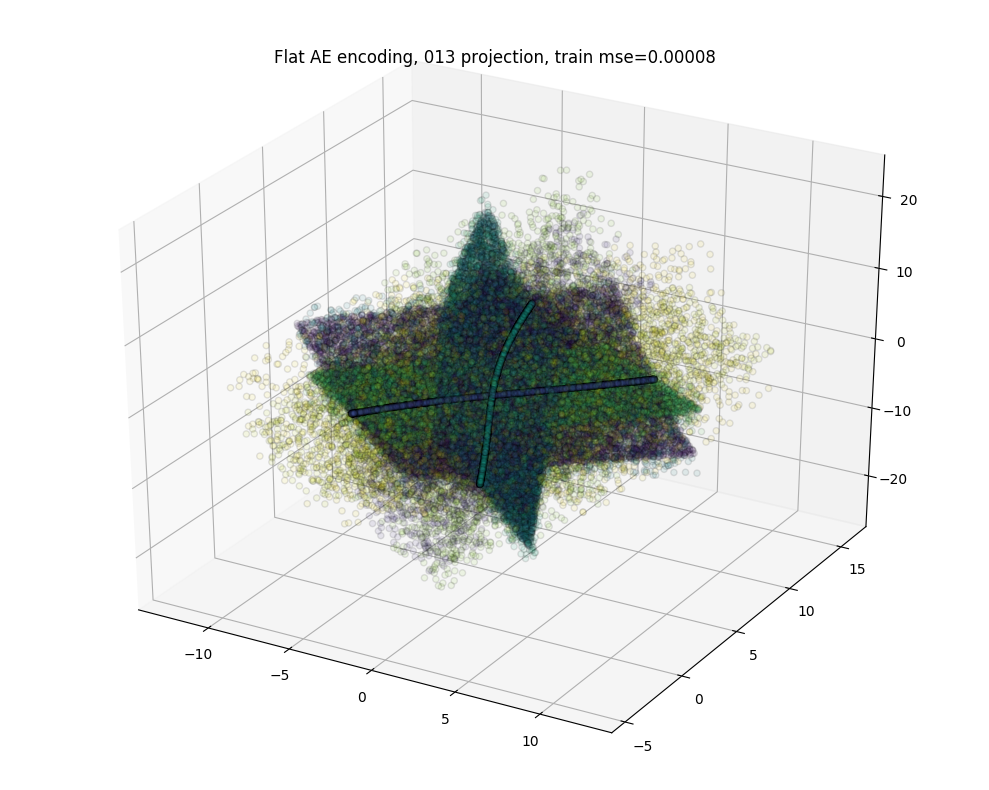}
\includegraphics[height=3cm]{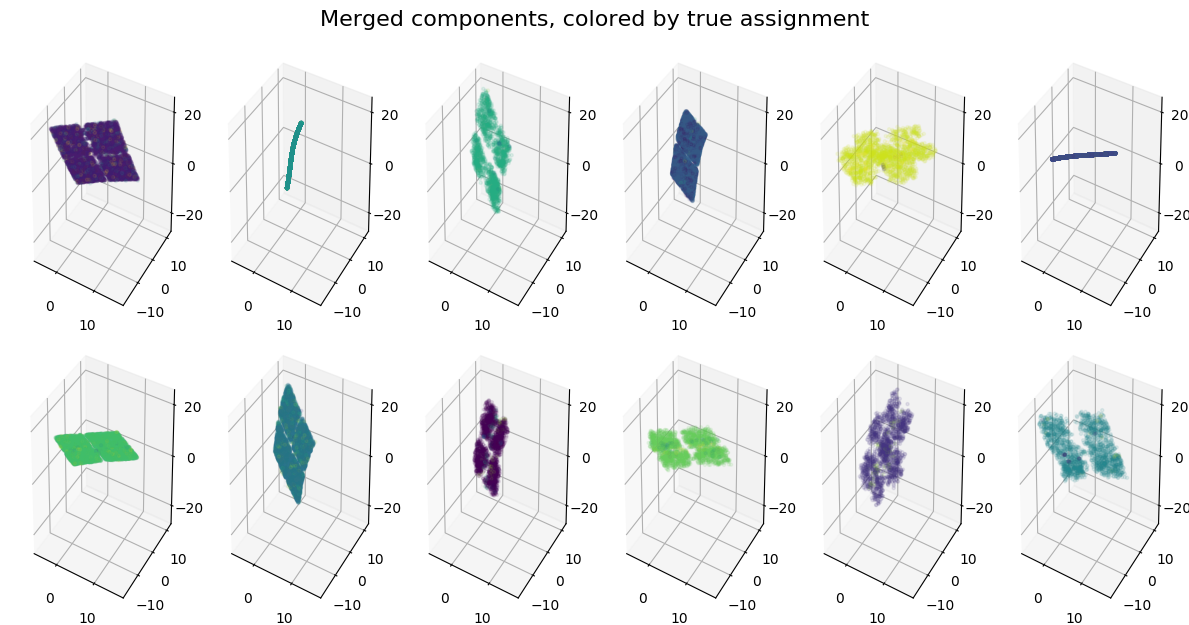}
\includegraphics[height=3cm]{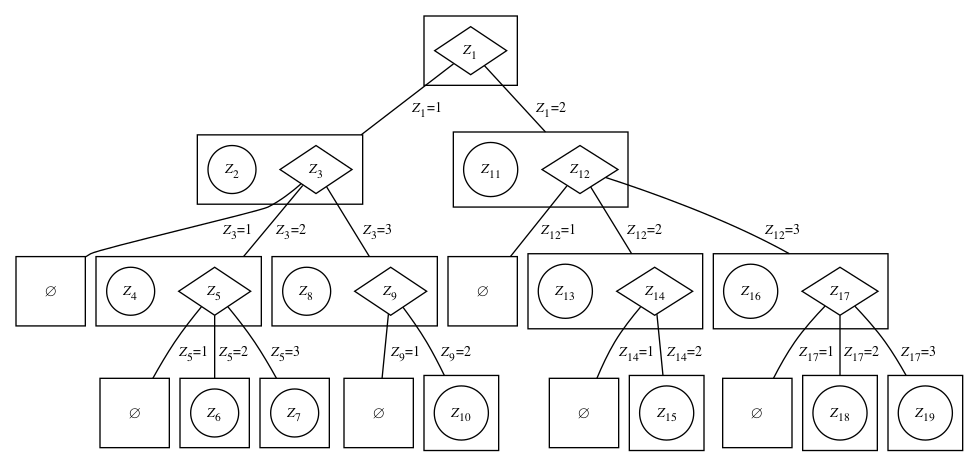}}
\caption{MIMOSA-learned initial encoding (left), 1D-3D components (middle), and hierarchy (right) for depth-3 Chopsticks varying either slope or intercept. Note that the learned hierarchy is not quite correct (two nodes at the deepest level are missing). Initial points are in 5 dimensions and projected to 3D. Analogue of Fig.~\ref{fig:chopsticks-manifolds}.}
\label{fig:chopsticks322-manifolds}
\end{figure*}

\begin{figure*}
    \centering

    \includegraphics[width=\textwidth]{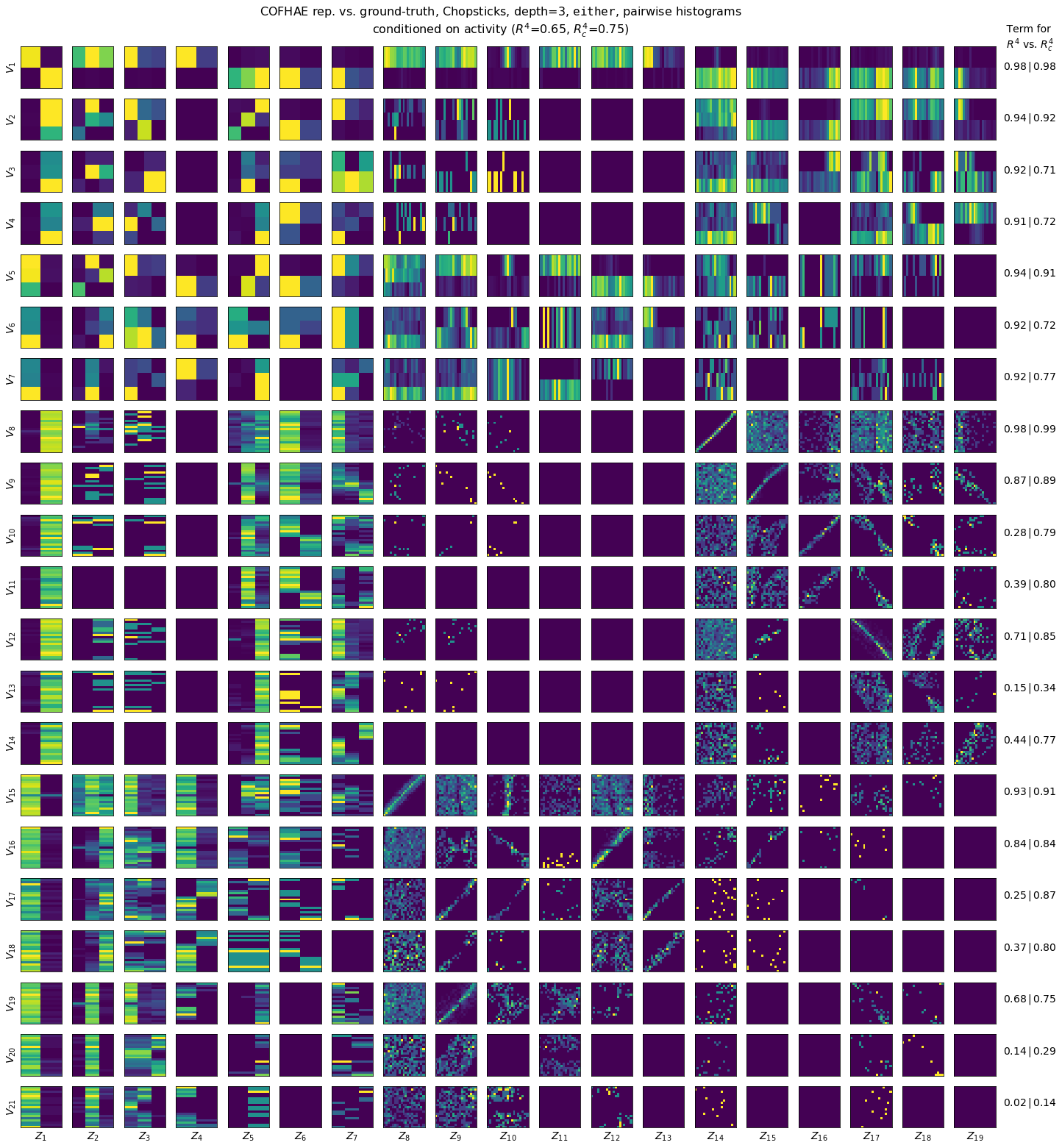}
    \caption{Pairwise histograms of ground-truth vs. learned variables for COFHAE on the most complicated hierarchical benchmark (Chopsticks at a recursion depth of 3 varying $\mathtt{either}$ slope or intercept). Histograms are conditioned on both variables being active, and dimension-wise components of the $R^4_c$ score are shown on the right.  Despite the depth of the hierarchy, COFHAE representations model it fairly well.}
    \label{fig:chopsticks-322-qual}
\end{figure*}

\end{document}